\documentclass{article}

\usepackage{arxiv}

\usepackage[utf8]{inputenc} 
\usepackage[T1]{fontenc}    
\usepackage{hyperref}       
\usepackage{url}            
\usepackage{booktabs}       
\usepackage{amsfonts}       
\usepackage{nicefrac}       
\usepackage{microtype}      
\usepackage[numbers]{natbib}
\usepackage{amsmath}
\usepackage{cleveref}       
\usepackage{lipsum}         
\usepackage{graphicx}
\usepackage{doi}

\usepackage{adjustbox}
\usepackage{amsmath}
\usepackage{array}
\usepackage{csquotes}
\usepackage{booktabs}
\usepackage{subcaption}
\usepackage{tabularx}

\title{I Am \textsc{AdMan}: A Pipeline for Automatic Generation of Personalized Advertising Imagery}

\date{}

\newif\ifuniqueAffiliation

\ifuniqueAffiliation 

\else
\usepackage{authblk}

\author[1,2]{Victor Kolominsky-Rabas\thanks{\texttt{victor.kolominsky-rabas@fit.fraunhofer.de}}}
\author[1,2]{Leopold Müller}
\author[1]{Claudius Budcke}
\author[1,2]{Niklas Kühl}

\affil[1]{University of Bayreuth, Universitätsstraße 30, 95447 Bayreuth, Germany}
\affil[2]{Fraunhofer FIT, Wittelsbacherring 10, 95444 Bayreuth, Germany}
\fi

\renewcommand{\headeright}{PREPRINT}
\renewcommand{\undertitle}{PREPRINT}
\renewcommand{\shorttitle}{I Am \textsc{AdMan}: A Pipeline for Automatic Generation of Personalized Advertising Imagery}

\hypersetup{
pdftitle={I Am \textsc{AdMan}: A Pipeline for Automatic Generation of Personalized Advertising Imagery},
pdfauthor={Victor Kolominsky-Rabas},
}

\begin{document}
\maketitle

\begin{abstract}
Personalized marketing can increase customer engagement, satisfaction, and conversion. While existing personalization approaches have become effective at matching the right product to the right customer, the visual representation of advertisements remains generic and only weakly tailored to the individual. Prior research shows that generative artificial intelligence can improve the creation of personalized advertisements, particularly for text, and that image generation models can support scalable advertisement production. However, little research has examined how detailed customer information can be systematically translated into fully AI-generated, personalized advertising imagery at scale on a technical level. To address this gap, we propose \textsc{AdMan}, a multi-agent pipeline that transforms customer data into personas, generates personalized advertisement images conditioned on product reference images, and applies an LLM-based judge agent for automated quality control. We implement the pipeline with two different model configurations and evaluate it across four products, using six celebrity personas for qualitative inspection, and $100$ real customer profiles, producing $1745$ advertisements. The evaluation combines a qualitative expert focus group and a quantitative artifact-rate assessment. The results show that the pipeline can generate photorealistic and personalized advertisements. At the same time, performance varies substantially by product complexity and model configuration. Our findings extend the literature on AI-based personalized advertising by demonstrating the feasibility and current limitations of fully automated image generation for advertising.
\end{abstract}



\section{Introduction}
Personalized marketing represents a critical advancement in contemporary digital marketing strategies, directly influencing customer engagement, satisfaction, and conversion rates~\citep{chandra2022personalization}.
Most current data-driven personalization approaches focus on matching the right product to the right customer, but only superficially consider presenting the product in a way that resonates with the individual customer~\citep{chandra2022personalization,murray2009personalization,naumov2019deep}. 
When it comes to advertising content, businesses have long relied on segmentation to create different advertisements for different target groups. Advertisements are still typically created manually and in advance and are generalized for a given target group, limiting their degree of personalization~\citep{smolinski2023towards}.
As competition intensifies and customer expectations rise, moving beyond targeting and product recommendation toward personalized presentation of advertisements could prove to be a significant competitive edge.
However, current marketing practices have not yet fully embraced this potential.
While targeting strategies have become increasingly precise, the visual and textual representation of advertisements often remains generic.
Recent advancements in generative artificial intelligence (GenAI), particularly large language models (LLMs), open new opportunities for hyper-personalized marketing content generation~\citep{lee2024developing,matz2024potential,patil2024generative}.
These technologies offer the potential to move beyond pre-designed content and can produce personalized advertisements at scale.
Parallel lines of research in AI and marketing automation emphasize the utility of LLMs for textual content generation \citep{lee2024developing,matz2024potential} and the creative potential of diffusion-based image generation and explore how generative AI can enhance visual appeal and scalability in advertisement creation~\citep{vashishtha2024chaining,czapp2024dynamic,yang2024new,hartmann2025power}.
A key challenge lies in determining how to systematically translate customer data---capturing demographics, interests, behaviors, and values---into creative content that is tailored not just in substance but also in style and presentation---and all at scale.
Yet, while personalized marketing and generative content creation are both well-studied domains, there is limited research on combining advancements from both fields for advertising imagery. Existing work in this space either focuses on narrow tasks like background inpainting rather than full advertisement generation~\citep{czapp2024dynamic,ku2023staging, shilova2023adbooster,yang2024new}, or achieves only coarse personalization by considering few dimensions of customer data
~\citep{smolinski2023towards,vashishtha2024chaining,xu2025personalized}, rather than systematically leveraging extensive profiles encompassing demographics, values, and lifestyle. No existing study addresses the automatic generation of personalized image-based advertisements at scale using detailed individual customer profiles. To address this research gap, we explore the following research questions:

\begin{itemize}
    \item [RQ1] How can we effectively utilize generative artificial intelligence for the fully automated generation of hyper-personalized advertising images?
    \item [RQ2] How does AI-based supervision improve image quality and robustness of the generation pipeline?
\end{itemize}

This work investigates these questions by designing and evaluating a prototype for such a system. Specifically, we propose and implement a fully autonomous pipeline that takes customer data as input and generates personalized advertising images.
We evaluate the pipeline using two model configurations: an OpenAI GPT configuration and a Google Gemini configuration. Across both configurations, we produce $1745$ advertising images using data of $100$ real customers and six celebrities.
We combine a qualitative expert evaluation with a quantitative analysis to assess the strengths and weaknesses of the pipeline, compare configuration-specific behavior, and derive future research avenues.
We want to to emphasize that the focus of this study is technical feasibility, not the effects of AI-based personalization on direct customer engagement. This topic is addressed in a separate study, which is outside the scope of this work.

Our results show that the pipeline produces personalized advertisements in which persona attributes such as hobbies, cultural background, and lifestyle cues propagate into the generated scenes. However, performance varies substantially by product complexity and model configuration. Products with simple visual structure achieve high quality, while products with prominent packaging text, complex object interactions, or intricate spatial arrangements remain challenging. Text rendering on products emerges as a persistent artifact source across model configurations. The Google Gemini configuration produces fewer human-labeled artifacts overall, but its judge often rejects images for minor details that are not relevant or visible to human reviewers. The OpenAI GPT judge, in contrast, can miss obvious physical artifacts. These findings delineate both the feasibility and the current boundaries of fully automated, persona-driven advertising image generation.

The remainder of this paper is organized as follows. \Cref{sec:background} reviews the relevant literature on personalized marketing and GenAI, establishing the theoretical foundation and related work. \Cref{sec:adman} introduces the \textsc{AdMan} pipeline. \Cref{sec:experimental-setup} describes the implementation, two model configurations, data collection, and evaluation procedure. The results of the qualitative evaluation and the quantitative comparison of the OpenAI GPT and Google Gemini configurations are reported in \Cref{sec:results}. Finally, \Cref{sec:discussion} discusses and interprets the results, addresses theoretical and practical implications and limitations, and suggests directions for future research. \Cref{sec:conclusion} concludes the work.
\section{Background \& related work}
\label{sec:background}
This section positions our work at the intersection of (i) personalization in marketing, (ii) generative AI for personalized content generation, and (iii) LLM-based multi-agent systems for scalable creative workflows.

\subsection{Generative artificial intelligence}
\label{subsec:genai}
The term AI refers to computer systems or machines that understand data, learn from it, and flexibly adjust to reach specific goals~\citep{haenlein2019brief}. Beyond this, GenAI produces novel content---text, images, audio, video---based on patterns learned from data~\citep{feuerriegel2024generative}. LLMs such as GPT-3~\citep{brown2020language} demonstrated that scaling transformer-based architectures enables strong few-shot and zero-shot capabilities for text generation. In parallel, diffusion-based image generation models such as Stable Diffusion~\citep{rombach2022high} and DALL$\cdot$E~\citep{ramesh2021zero,betker2023improving} can synthesize high-fidelity images from text prompts, supporting detailed control over style and composition. Together, these advances provide the technological foundation for multimodal content generation pipelines in which text models produce structured creative instructions and image models render corresponding visuals.

\subsection{Multi-agent systems}
\label{subsec:mas_marketing}

Multi-agent systems (MAS) comprise multiple autonomous agents that perceive their environment, communicate, and coordinate actions to solve tasks that exceed the capabilities of a single agent~\citep{allmendinger2026multi,wooldridge2009introduction}. With the rise of LLMs, MAS research has increasingly explored language-mediated agentic frameworks in which role-specialized LLM agents collaborate, critique, and iteratively refine outputs~\citep{tran2025multi,yang2024llm,han2024llm,acharya2025agentic}. Such systems are often motivated by the observation that decomposing tasks across specialized agents can mitigate limitations of single LLMs (e.g., brittle reasoning, hallucination, and limited self-critique) through structured interaction and verification~\citep{tran2025multi}. When combined with tools, planning, and external knowledge sources, LLM-based MAS can support complex, multi-step workflows in a scalable manner~\citep{talebirad2023multi,tran2025multi}.

These properties make LLM-based MAS a natural fit for marketing tasks that require coordinated analysis, content generation, and evaluation. For instance, \citet{xiao2024tradingagents} illustrate the benefits of role specialization in a stock-trading setting through agents that emulate organizational functions. In marketing, \citet{quan2025crmagent} propose CRMAgent, which coordinates agents for analysis, retrieval, generation, and evaluation to rewrite underperforming campaign messages. Beyond text, multi-agent frameworks such as CREA show how role-based agents can collaboratively generate and critique images, improving novelty and alignment with creative goals~\citep{venkatesh2025crea}. Despite these advances, prior research has not investigated how LLM-based MAS can be used to translate detailed customer persona descriptions into tailored prompts that drive personalized image-based advertisement generation at scale---a gap we address with the pipeline proposed and evaluated in this work.

\subsection{Personalization in marketing}
\label{subsec:image_marketing_personalization}
Personalization refers to the adjustment of a product, a service, or even an experience to match an individual's characteristics, preferences, and/or needs~\citep{cavdar2021typology,peppers1993one}. If applied to the field of marketing, the meaning shifts and becomes more embedded in the context of the marketing mix: product, price, place, promotion. The information about a customer's profile is utilized to alter any component of the marketing mix in order to gain a competitive advantage~\citep{chandra2022personalization,montgomery2009prospects}.
Usually, this is achieved by matching an offering to the customer's profile, thereby increasing the relevance of that offering for the customer and creating a curated and enhanced experience~\citep{polk2020magic}. The advantage of personalized offerings stems from a boost in customer attention~\citep{bang2016tracking}, which translates to higher engagement and conversion rates~\citep{ag2024personalized,chandra2022personalization}.

In practice, personalized marketing involves the targeting of individuals with advertisements and recommending certain products or services. Based on what is known about a customer's profile, e.g., their level of income or their interests and hobbies, it is possible to determine which products are relevant to their everyday lives and hence make a recommendation on what to buy next. This process is known as personalized recommendation~\citep{chandra2022personalization}.
At the same time, it is possible to adjust the contents of advertisements to better match customer tastes or increase recognizability. Athletes promote products to sports fans, while actors and actresses advertise products to movie-enthusiasts. The adjustment of advertising contents is called personalized advertising~\citep{chandra2022personalization}.

\subsection{AI in personalized marketing}
\label{subsec:ai_marketing_personalization}
These activities in personalized marketing are greatly enhanced by AI and the capabilities for processing and analyzing vast amounts of data it provides. For example, AI is able to analyze browsing histories and purchasing patterns as well as customer sentiment to determine the likes and dislikes of customers. Combined with content and demographics from social media, holistic customer profiles can be created on an individual level and cross-referenced with other users with shared interests~\citep{aivalis2016evolving}. Nowadays, AI is ubiquitous in the targeting of customers through recommendation systems and is widely studied and utilized~\citep{chandra2022personalization,dritsas2025machine,huang2025recommender}. Besides the success of AI in personalized recommendations, it has also led to advancements in personalized advertising. For example, \citet{li2024two} design an AI-based system that dynamically chooses advertisement imagery from a pool of available material based on user preferences. While the creation of advertisement material has been a manual and expensive process in the past~\citep{li2024two,matz2024potential}, GenAI now makes it possible to adjust texts, images or even video dynamically, faster, and with comparably low cost~\citep{kapoor2025frontiers,matz2024potential,noy2023experimental}. Not only is AI-generated content cheaper, but under certain circumstances, it has also been shown to perform better than human-generated content and is perceived as more effective and more substantiated than human-authored content~\citep{karinshak2023working,matz2024potential,reisenbichler2026applying}. Similarly, ~\citet{zhang2022automatic} report that AI-generated product descriptions increase click-through rate and conversion rate.

These achievements are not only limited to textual generation.~\citet{hartmann2025power} show that AI-generated imagery is not only able to compete with human-generated advertisements, but can increase click-through rate significantly. Driven by this potential of GenAI for advertisement creation, a growing body of work investigates generative approaches for producing personalized marketing images~\citep{chen2025ctr,czapp2024dynamic,mishra2020learning,shilova2023adbooster,wang2025generate,yang2024new}. This work, where models generate or adapt advertisement content to optimize product presentation, aesthetics, and context, based on engagement metrics like the click-through rate, is also termed generative creative optimization~\citep{shilova2023adbooster}. Many of these approaches use inpainting, where they take a product image and generate new backgrounds to match customer preferences~\citep{czapp2024dynamic,ku2023staging,shilova2023adbooster,yang2024new}. However, these approaches do not consider the full generation of advertising content, based on extensive and detailed customer profiles. Frameworks like Pigeon~\citep{xu2025personalized} utilize GenAI to create the entire advertisement imagery. Customer preferences are inferred from historic interaction with other images, which are processed by AI together with a reference image, to create the new personalized advertisement. However, the approach is limited by the exclusion of customer data beyond historic interaction and the narrow focus on movie posters in the study, which gives no indication of how the framework would perform with less homogeneous input data. Other approaches that consider fully AI-generated marketing imagery are also constrained in that they take customer preferences into account only to a limited extent. Oftentimes, only one or a few dimensions of personalization are considered, which simplifies image generation but also does not achieve true personalization on an individual level~\citep{smolinski2023towards,vashishtha2024chaining}.
Although research has made huge advancements in personalized content generation based on AI, existing work in personalized image generation still relies on narrow problems or coarser personalization and does not systematically utilize extensive and detailed customer information, such as demographics, values, lifestyle, in addition to preferences.
\section{\textsc{AdMan}: A pipeline for automatic generation of personalized advertising imagery}
\label{sec:adman}

\begin{figure*}[tb]
	\centering
	\includegraphics[width=0.8\textwidth]{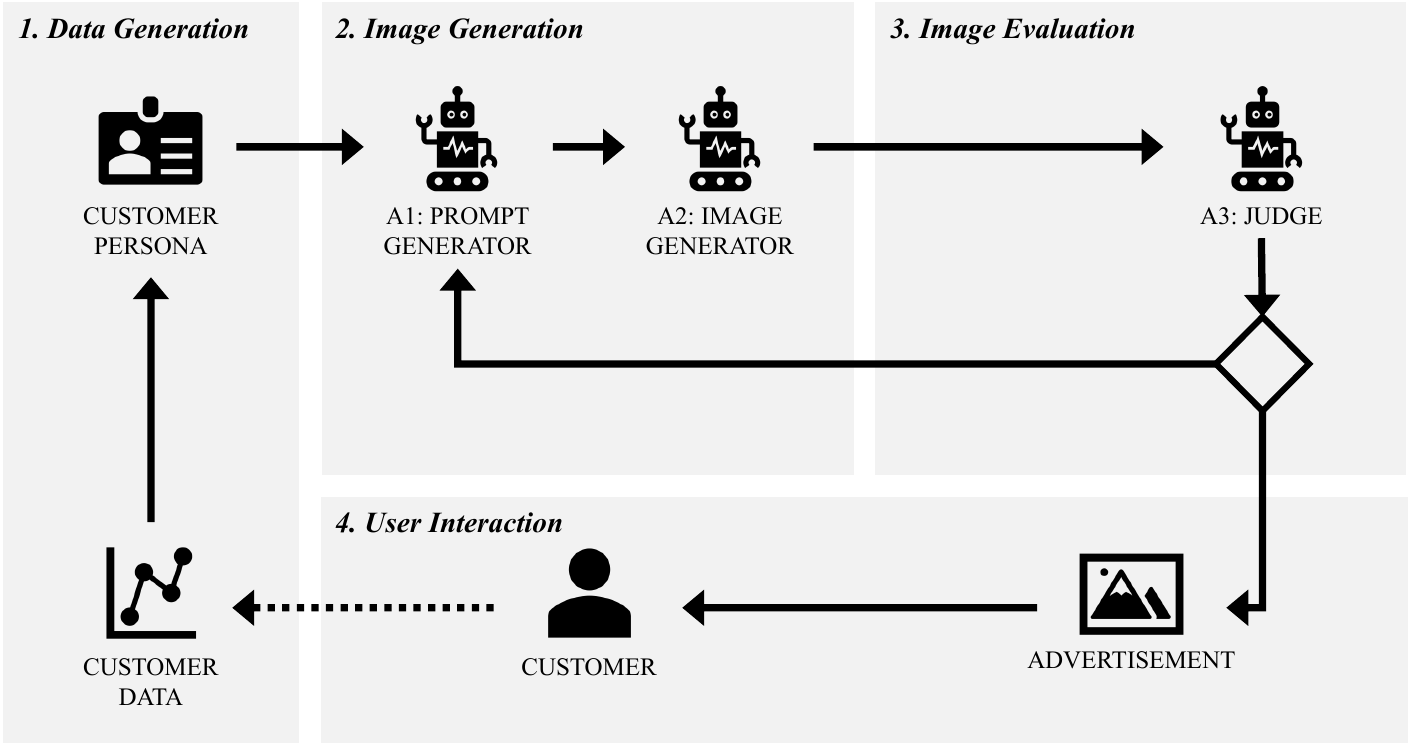}
	\caption{\textsc{AdMan}: a multi-agent pipeline that converts customer data into a persona, generates a personalized advertising image conditioned on a product reference image, evaluates the generated outputs, and serves a final image to the customer.}
	\label{fig:adman}
\end{figure*}

We investigate how personalized advertising images can be generated fully automated, based on data from individual customers, using GenAI. We propose \textsc{AdMan}, a personalized \textbf{ad}vertising image \textbf{M}ulti-\textbf{a}gent-based ge\textbf{n}eration pipeline. 
\textsc{AdMan} is inspired by established advertisement creation workflows, but replaces manual creative work with a structured, agentic pipeline. It consists of four stages: (1) data generation, (2) image generation, (3) image evaluation, and (4) user interaction. \Cref{fig:adman} summarizes the pipeline.

\paragraph{Data generation.}
The pipeline starts once a product has been selected for display to a specific customer (e.g., after eligibility and targeting decisions in an advertisement delivery system). Unlike conventional pipelines that match a customer to a pre-created advertisement, \textsc{AdMan} uses customer information to personalize the advertisement itself. We assume access to customer attributes that are commonly collected or inferred in digital advertising ecosystems (e.g., demographics, interests, and behavioral signals). These attributes are transformed into a \emph{customer persona}. In our conceptualization, the persona is a textual description, which can be directly consumed by LLMs and text-to-image models. In future applications, personas could alternatively be represented as embeddings that capture richer, continuously updatable user information. This would require models capable of conditioning on such representations, but could align well with standard representations already used in marketing pipelines. The persona and a product reference image serve as inputs for the next stage.

\paragraph{Image generation.}
The image generation stage involves two agents. The \emph{prompt generation agent} (A1) takes the customer persona as input and produces an image-generation prompt that specifies a plausible, appealing scene composition and style aligned with the persona. This prompt, together with the product reference image, is provided to the \emph{image generator agent} (A2). The goal of A2 is to generate an advertisement image that (i) reflects the persona through the depicted context, aesthetics, and atmosphere, and (ii) presents the product in a visually coherent manner consistent with the reference image. This stage offers multiple degrees of freedom for design, e.g., underlying models, prompt templates, product- or brand-specific constraints, and additional control layers for policy and content compliance. In this work, we focus on a single proof-of-concept instantiation and leave systematic comparisons of alternative design choices for future research.

\paragraph{Image evaluation.}
The third stage evaluates generated outputs before delivery. A \emph{judge agent} (A3) assesses whether an image meets minimum quality standards and whether it contains obvious artifacts or implausible elements that would undermine its usability as an advertisement. Design choices include the information available to the judge (e.g., generated image only vs.\ generated image plus product reference), the assessment format (binary pass/fail vs.\ graded scoring), and the remediation strategy when an image fails. If an image does not pass, \textsc{AdMan} can either (i) restart generation with revised prompts, (ii) perform iterative improvement using the judge feedback, or (iii) fall back to a default advertisement. In our proof-of-concept instantiation, we implement a simple regeneration strategy with a fixed version limit and a final best-of selection if no version passes.

\paragraph{User interaction.}
In the final stage, the resulting advertisement imagery is served to the customer. Depending on the chosen policy, this can be the first passing version, the highest-scoring version, the final version, or a fallback advertisement. In our instantiation, if no version passes, the judge agent (A3) selects the best available version among the generated candidates.
\section{Experimental setup}
\label{sec:experimental-setup}

First, we implement \textsc{AdMan}, which is laid out in~\Cref{subsec:implementation}. In \Cref{subsec:data-collection}, we describe the two types of customer data used for evaluation. The evaluation procedure is described in \Cref{subsec:evaluation} and consists of a qualitative and quantitative analysis. First, we use the OpenAI GPT configuration and celebrity customer data to conduct an initial qualitative evaluation, focusing on general feasibility, generation challenges, and whether the pipeline enables recognizable personalization. Second, we apply both model configurations (OpenAI GPT and Google Gemini) to customer data of $100$ real individuals and evaluate image quality regarding artifact and pass rate.

\subsection{Pipeline implementation}
\label{subsec:implementation}

To enable an evaluation, we implement \textsc{AdMan} with two fixed model configurations, shared prompt templates, and a bounded regeneration policy. The OpenAI GPT configuration uses OpenAI models for prompt generation, image generation, and judging. The Google Gemini configuration replaces these models with the corresponding Gemini models while keeping the remaining pipeline logic unchanged. An overview of the two model configurations can be seen in \Cref{tab:model-configurations}.\footnote{Code and dataset will be made publicly available upon publication.}

\paragraph{Customer data.}
Customer data is collected as tabular variables with predefined categories. We include variables based on three criteria: (1) relevance for image generation, (2) relevance for market segmentation (with hyper-targeting understood as segmentation at a higher level of granularity), and (3) plausibility of being collectible in practice (e.g., through data commonly available on major digital platforms).

We collect data from five categories: \textbf{(1) Basic information} (gender, age, country of residency, city of residency, country of birth, ethnicity, family origin/background) captures demographic and geographic inputs commonly used for segmentation and plausibly collectible in practice~\citep{kotlerMarketingManagement2016, googlePrivacyPolicyPrivacy2025}.
\textbf{(2) Life stage} (occupation type, occupation title, income class, relationship status, number of children, age groups of children) reflects established segmentation criteria and is plausibly inferable from behavior in practice~\citep{kotlerMarketingManagement2016, googlePrivacyPolicyPrivacy2025}.
\textbf{(3) Lifestyle} (regular activities, hobbies) captures psychographic signals to enable realistic, scenario-based depictions of product use; such signals can be inferred from consumption and search behavior~\citep{googlePrivacyPolicyPrivacy2025}.
\textbf{(4) Environment} (housing type, residence area, climate, attention check, pet ownership, and pet type) provides household and contextual variables relevant for coherent scene construction. Dwelling and area characteristics are recognized segmentation dimensions~\citep{wilkieConsumerBehavior1994, czinkota2021market}. Climate supports visually coherent settings. Pet ownership is prevalent and salient for lifestyle portrayal~\citep{shahbandehNumberPetOwning2025}. Residence categories follow Statistics Canada~\citep{statisticscanadaTypeDwellingReference2022}.
\textbf{(5) Style} (hair style, hair color, beard style, body type, glasses wearer, color preference, style preference) captures appearance and aesthetic preferences that directly inform visual personalization. Such attributes can be inferred from user images using human attribute recognition~\citep{li2016human}. This category may be particularly relevant for social media platforms, which have access to large amounts of user data, especially image data, from which visual customer information can be derived.

\paragraph{Customer persona.}
We transform tabular customer data into a persona description using a deterministic, template-based script that maps categorical values to a short textual profile. For the illustrative example \textit{Bruce Wayne} (The Batman movie), this yields:
\begin{quote}
\small
\textit{``The persona is a male individual, 39 years old, born in USA, currently living in Gotham, USA. He lives in a mansion, in an area that can be described as ``rich neighborhood''. The climate there is currently cloudy. He would describe his current occupation as ``billionaire''. His most recent occupation title is ``entrepreneur'', and he earns an extremely high income. He regularly engages in the following activities: working, parties, social gatherings. He is single. His appearance includes a slick back hairstyle, dark brown hair color, and a(n) very muscular build. He prefers dark blue colors and likes to dress in classic style.''}
\end{quote}

\paragraph{A1: Prompt generation agent.}
The prompt generation agent (A1) is implemented using gpt-5 in the OpenAI GPT configuration and gemini-3.1-pro-preview in the Google Gemini configuration. A1 is configured with a fixed system prompt that defines its role and expected output format. Its user prompt follows a generic template shared across all products and customers and is populated at runtime with the customer persona, product name, and product information. The system and user prompts are refined using prompt engineering techniques~\citep{liu2023pre,muller2025data}.

\paragraph{A2: Image generator agent.}
The image generator agent (A2) is implemented using gpt-image-1 in the OpenAI GPT configuration and gemini-3-pro-image-preview in the Google Gemini configuration. As the image generation model does not support system prompts, A2 receives only a user prompt. The user prompt template is identical across customers and products and is populated at runtime with the product name and the image-generation prompt produced by A1. In addition, A2 receives the product reference image to support faithful product placement and appearance.

\paragraph{A3: Judge agent and re-generation policy.}
The judge agent (A3) is implemented using gpt-5 in the OpenAI GPT configuration and gemini-3.1-pro-preview in the Google Gemini configuration. It serves two functions. First, it assesses whether a generated image contains conspicuous artifacts. For this check, A3 receives both the generated image and the product reference image. If the image passes, it is accepted as the final output. If the image fails, A3 triggers a regeneration cycle (A1 $\rightarrow$ A2). We set a version limit of three to reflect practical constraints. Second, if no version passes within the limit, A3 receives all previous judgments and selects the best available version as the final advertisement. Both judge functions use fixed prompt templates that are shared across products and customers and populated at runtime with product-specific information.

\paragraph{Products.}
We select four products to cover diverse advertising contexts while maintaining theoretical grounding. Product selection follows the FCB Grid \citep{vaughn1980advertising}, which classifies products by purchase involvement (low vs.\ high) and dominant decision driver (thinking vs.\ feeling)~\citep{vaughn1980advertising}. We select one product per quadrant to ensure conceptual diversity and to facilitate replication. We also consider expected visual portrayals (e.g., indoor vs.\ outdoor scenes) to avoid systematically biasing generated settings.

For the \textit{informative} quadrant, we select an electric vehicle~\cite{polestar2024}, which typically involves high involvement and largely rational decision-making~\citep{cheong2017revisiting}. This choice is inspired by \citet{jansen2023automated}. For the \textit{affective} quadrant, we select a luxury watch~\citep{rolex2025}, representing emotionally driven, high-involvement luxury purchases~\citep{teng2010use}. For the \textit{habitual} quadrant, we select laundry detergent~\citep{tide2020}, reflecting low-involvement, routine decision-making characteristic of everyday household products~\citep{glowa2002white} and for the \textit{self-satisfaction} quadrant, we select a soft drink~\citep{cola2025}, representing low-involvement and emotionally driven consumption decisions~\citep{cheong2017revisiting}.

\begin{table}[ht]
\centering
\begin{tabular}{lll}
\hline
\textbf{Configuration} & \textbf{OpenAI GPT} & \textbf{Google Gemini} \\
\hline
A1 & gpt-5 & gemini-3.1-pro-preview \\
A2 & gpt-image-1 & gemini-3-pro-image-preview \\
A3 & gpt-5 & gemini-3.1-pro-preview \\
\hline
\end{tabular}
\caption{Comparison of the two model configurations.}
\label{tab:model-configurations}
\end{table}

\subsection{Data collection}
\label{subsec:data-collection}

\paragraph{Celebrity customer data.}
To make personalization judgments intuitive, we use a set of celebrities as customers: their widely known characteristics (e.g., appearance, interests, and lifestyle cues) facilitate assessing whether generated outputs plausibly reflect the intended persona. Importantly, \textsc{AdMan} is not informed about the identity of the celebrity. It only receives the persona attributes. An exception is \textit{Bruce Wayne}, for whom the hometown ``Gotham'' is unique. We select the following six celebrities: Bruce Wayne (The Batman movie), Vito Corleone (The Godfather movie), Rajesh Koothrappali (The Big Bang Theory series), Rachel Green (Friends series), Morpheus (The Matrix movie), and Yan Naing Lee (Rush Hour movies).

\paragraph{Real customer data.}
To obtain a set of real customer profiles, we conduct a Prolific \citep{prolific} study with $N=100$ filtered participants who complete a survey covering all customer data variables. Participants have a mean age of $45.8$ years ($SD = 13.9$), ranging from 22 to 84 years. The gender distribution is balanced (47 male, 49 female, 4 diverse). Most participants report being born in the United States (91\%), consistent with the recruitment criteria. Ethnic background is reported as 81 White (81\%), 6 Asian (6\%), 5 Hispanic (5\%), 4 Black (4\%), and 4 Other (4\%). All participants are native English speakers. Regarding employment, 73 participants (73\%) report working (full-time, part-time, or self-employed), 13 (13\%) report being unemployed, 12 (12\%) are retired, and 2 (2\%) are university students. The median completion time is $4.3$ minutes ($M = 5.1$, $SD = 2.8$, range = $1.8$-$18.9$).
The Research Ethics Committee at University of Bayreuth approved our study (approval: 25-111) on July 09, 2025. Respondents gave written consent for review and signature before starting the survey.

\subsection{Evaluation}
\label{subsec:evaluation}
We evaluate the pipeline to assess feasibility, personalization, and key failure modes. The evaluation combines a qualitative expert focus group to surface strengths and challenges and a quantitative artifact-rate assessment to derive further insights. We use identical generation settings across all products and personas.

\subsubsection{Qualitative Evaluation}
\label{subsubsec:focus_group}

Our first evaluation component is a focus group with eight AI researchers, three of whom are part of the author team. The author team participants moderated the workshop and provided methodological context. The substantive assessment was based on the discussion and feedback of the non-author experts. In a two-hour workshop, the five non-author experts reviewed 24 generated advertisements (six celebrities $\times$ four products). For this initial evaluation, we only use the OpenAI GPT configuration because the objective is to evaluate the general feasibility of the pipeline, generation challenges, and whether personalization is generally achieved. All reviewed images are the final version (the intermediate versions are not considered in the qualitative evaluation), which has been selected by the judge. In addition to the images, the personas are shown to the experts. They spot and point out obvious mistakes, rate the model on consistency and quality, identify recurring problems, and discuss potential root causes of any issues that arise. Celebrities were chosen to create some familiarity with the persona descriptions and generated images. Instead of seeing descriptions and images of strangers, experts had some expectation of what images should look like---as they would for personalized advertisements---to increase the realism of the evaluation setting.
We code the statements and aggregate the qualitative observations to derive a set of strengths and challenges, as well as limitations and implications for future research.

\subsubsection{Quantitative Evaluation}
\label{subsubsec:artifact_rate}

Using the collected customer data, we run each model configuration to generate advertisements for each of the four products and each of the $100$ customers. This yields $4 \times 100 = 400$ final advertisements per configuration, or $800$ final advertisements across both configurations. Additional intermediate images are produced when regeneration is triggered. Their number depends on the judge decisions and therefore differs between configurations. We include these intermediate images when computing metrics for $I_{\text{first}}$ and $I_{\text{passed}}$, but we do not report the number of intermediate images separately. Subsequently, we review all generated images to create a set of ground-truth labels for them. Conflicts are resolved by discussion. We report all metrics per product and as an overall mean (\Cref{tab:quantitative_results}).

\paragraph{Image sets}
For each model configuration, we define and analyze three subsets of the set $I$ of all generated images:
\begin{itemize}
    \item $I_{\text{first}}$: The first-generated image for each customer--product pair ($|I_{\text{first}}| = 400$ per configuration). This set reflects baseline generation quality before any regeneration.
    \item $I_{\text{passed}}$: All images across all versions that passed the judge's evaluation. Its size depends on how many images were rated as pass or fail and therefore varies by product.
    \item $I_{\text{selected}}$: The final image delivered to the customer for each customer--product pair ($|I_{\text{selected}}| = 400$ per configuration). This is either the first passing version or, if no version passes within the version limit, the best candidate selected by the judge (A3).
\end{itemize}

\paragraph{Pass rate.}
The pass rate ($PR$) measures the share of images in a set approved by the judge. Formally, $PR = n_{\text{passed}} / n_{\text{total}}$, where $n_{\text{passed}}$ is the number of images that passed and $n_{\text{total}}$ is the set size. For $I_{\text{first}}$, $PR$ indicates how often the first generation attempt is accepted without regeneration. For $I_{\text{passed}}$, $PR$ is trivially $1.0$ by design. For $I_{\text{selected}}$, $PR$ indicates how often the final delivered image actually passed the judge (as opposed to being selected as the best available candidate after all versions failed).

\begin{figure*}[tb]
    \centering
    \small
    \setlength{\tabcolsep}{2pt}
    \renewcommand{\arraystretch}{1.15}

    \begin{tabular}{@{}>{\centering\arraybackslash}m{0.12\textwidth}
                    >{\centering\arraybackslash}m{0.205\textwidth}
                    >{\centering\arraybackslash}m{0.205\textwidth}
                    >{\centering\arraybackslash}m{0.205\textwidth}
                    >{\centering\arraybackslash}m{0.205\textwidth}@{}}
        \toprule
        & \textbf{Electric vehicle} & \textbf{Luxury watch} & \textbf{Laundry detergent} & \textbf{Soft drink} \\
        \midrule
        \shortstack{\textbf{OpenAI}\\\textbf{GPT}} &
        \includegraphics[width=\linewidth]{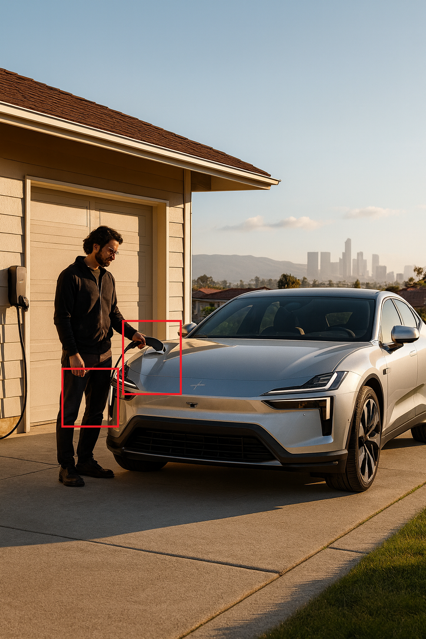} &
        \includegraphics[width=\linewidth]{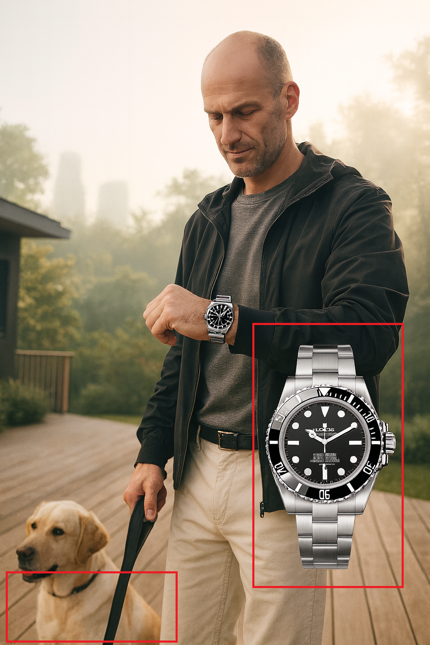} &
        \includegraphics[width=\linewidth]{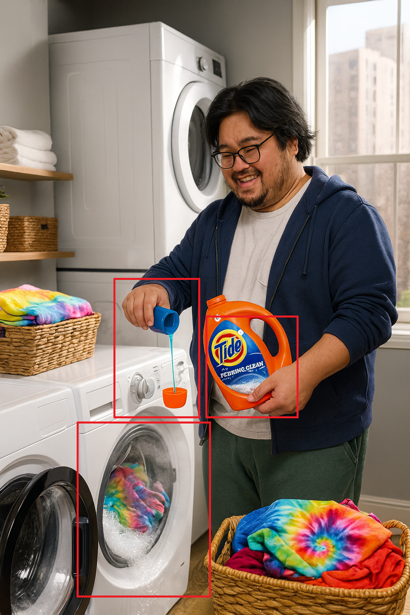} &
        \includegraphics[width=\linewidth]{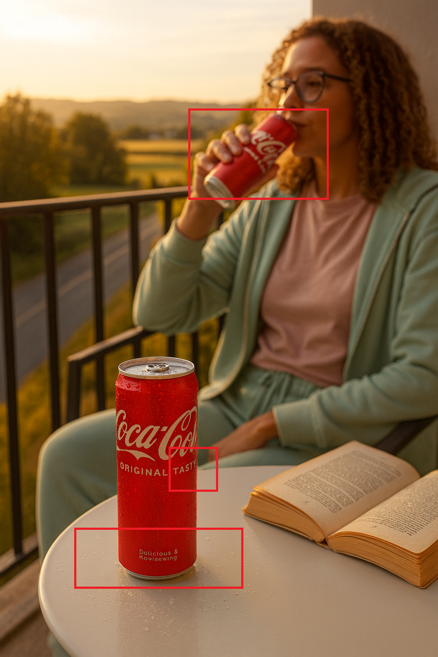} \\
        \addlinespace[4pt]
        \shortstack{\textbf{Google}\\\textbf{Gemini}} &
        \includegraphics[width=\linewidth]{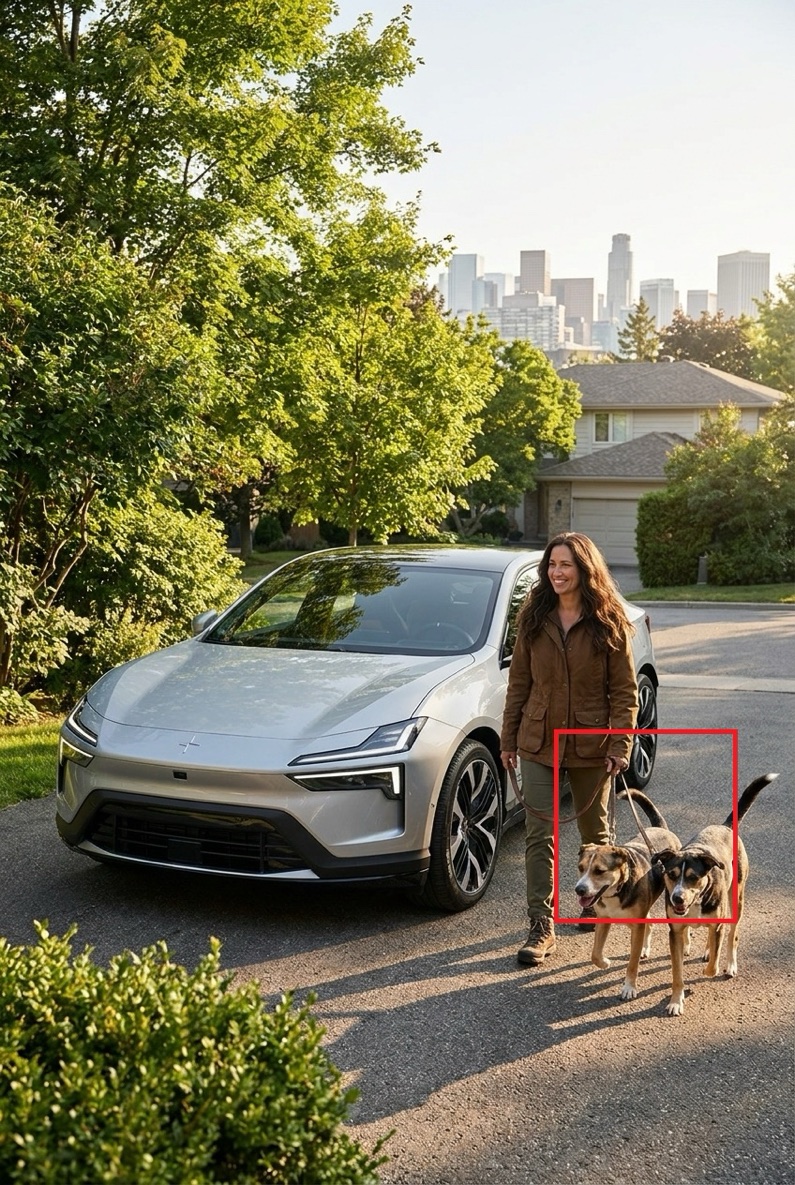} &
        \includegraphics[width=\linewidth]{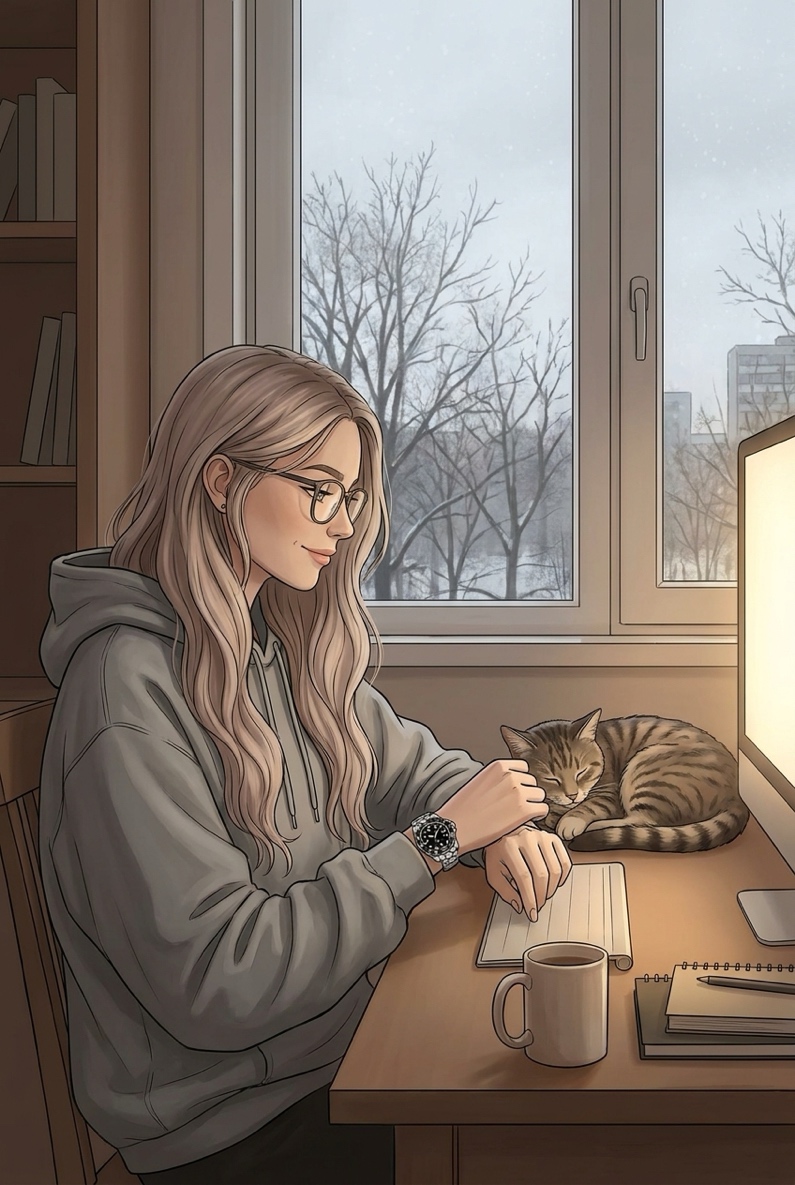} &
        \includegraphics[width=\linewidth]{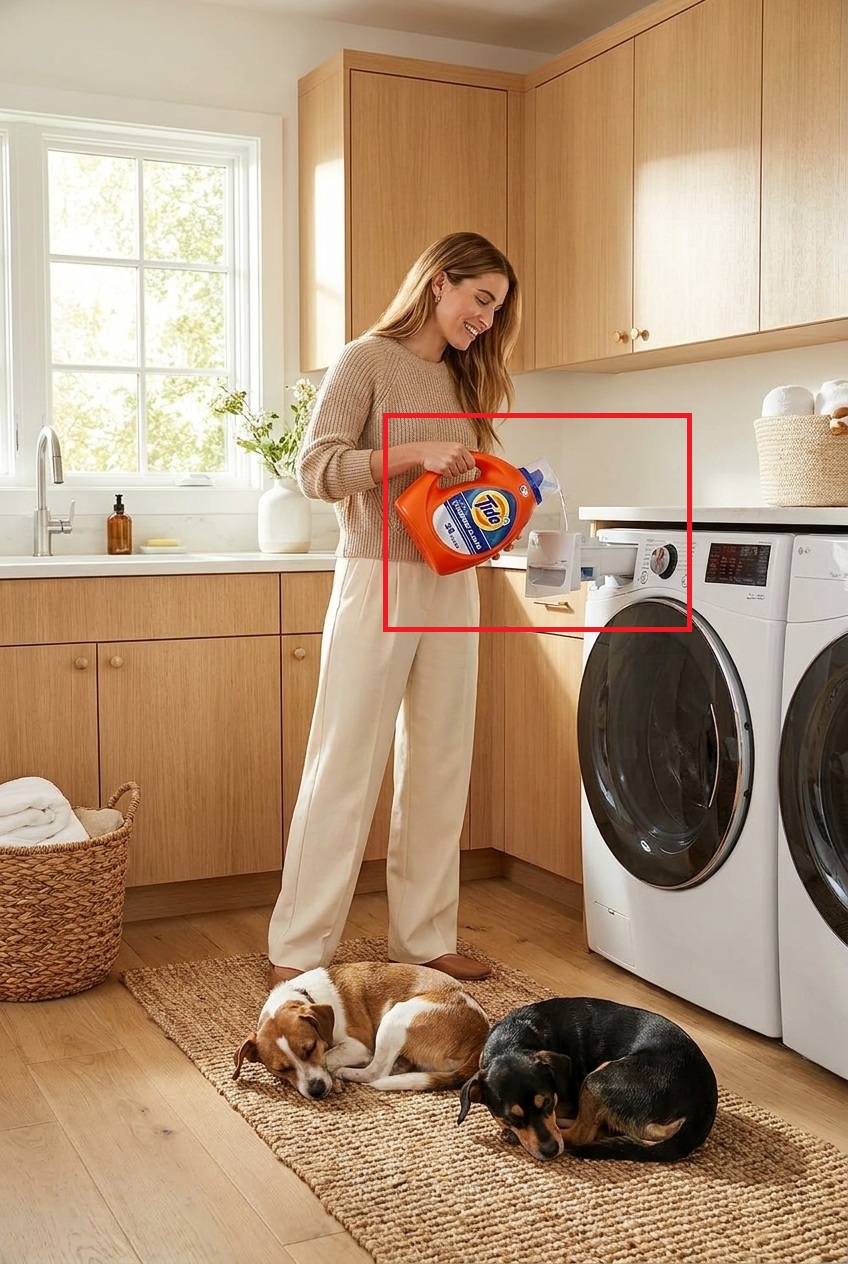} &
        \includegraphics[width=\linewidth]{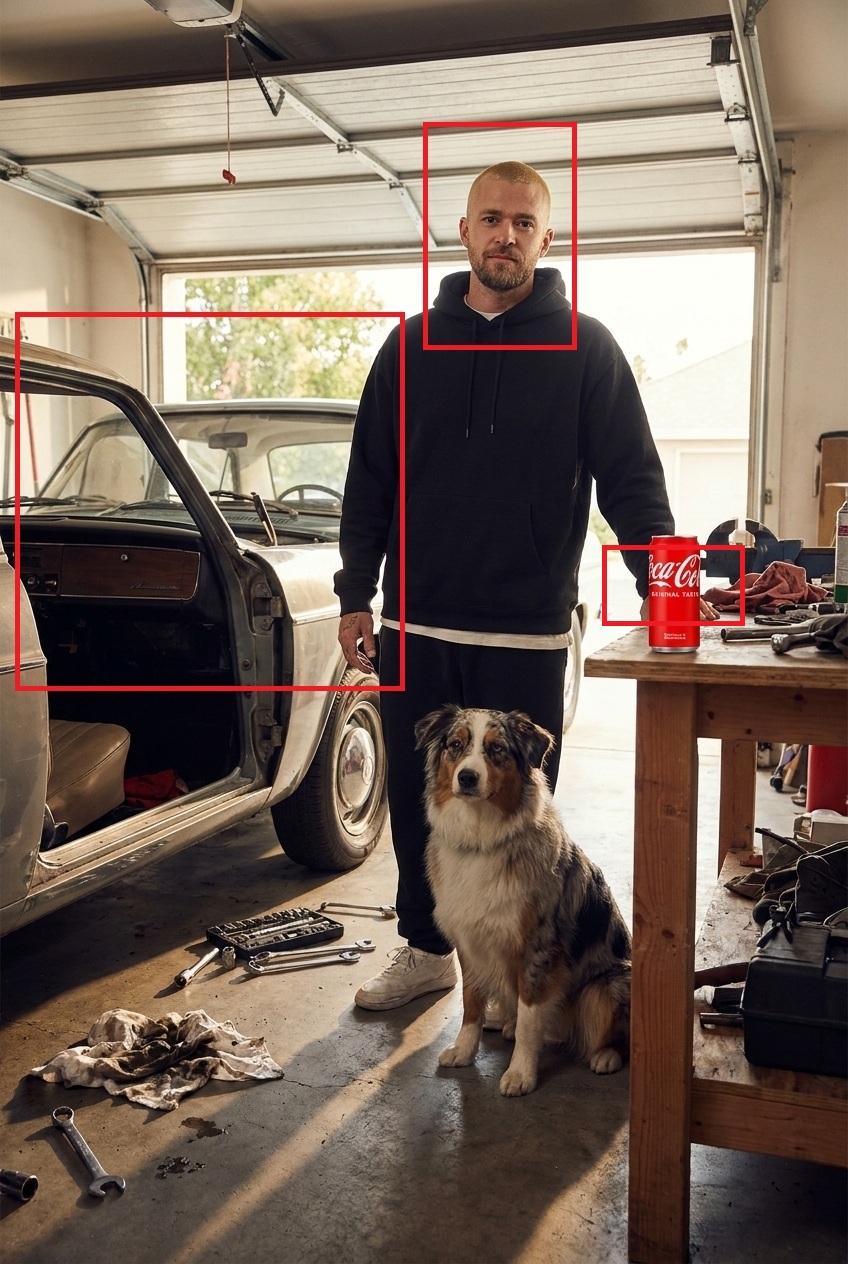} \\
        \bottomrule
    \end{tabular}
    
    \caption{Some images contain obvious artifacts across the OpenAI GPT and Google Gemini configurations. Common artifacts include erroneous docking for the electric vehicle, missing cables or leashes, prominent floating watches, missing doors on laundry machines, unrealistic image style or distorted and nonsensical text. Major artifacts are highlighted in red.}
    \label{fig:example_images_bad}
\end{figure*}

\paragraph{Artifact rate.}
We define an \emph{artifact} as an unrealistic defect in the image that would plausibly reduce its usability as an advertisement (e.g., breaking physical realism or missing structural elements such as a car door or a washing machine door). \Cref{fig:example_images_bad} shows exemplary artifacts in real image examples.

Each image (including intermediate versions) is independently reviewed by three reviewers.
The reviewers assign artifact labels based on visual inspection only. To avoid influencing the reviewers' decision-making process, the labeling decision does not use pass or fail decisions of the judge agent and version metadata.
A majority vote determines whether an image is labeled as containing an artifact.
The artifact rate $AR$ is the share of images in a set that contain at least one artifact: $AR = n_{\text{artifact}} / n_{\text{total}}$.
Because text rendering remains a known limitation of current image-generation models, we additionally report a relaxed artifact rate $rAR$ that excludes images only containing text-based artifacts. Comparing $AR$ and $rAR$ across $I_{\text{first}}$, $I_{\text{passed}}$, and $I_{\text{selected}}$ reveals (i) the baseline artifact prevalence, (ii) whether the judge effectively filters artifacts, and (iii) the artifact quality of the final deliverables.

\paragraph{Statistical validation.}
To quantify statistical uncertainty, we compute Wilson 95\% confidence intervals for all proportions ($PR$, $AR$, and $rAR$).
Because each participant contributes one image per product in $I_{\text{first}}$ and $I_{\text{selected}}$, product-level and configuration-level comparisons in these two sets are paired at the customer--product level.
We therefore use exact McNemar tests for pairwise comparisons in $I_{\text{first}}$ and $I_{\text{selected}}$.
For $I_{\text{passed}}$, set membership depends on the judge's pass/fail decisions and denominators differ across products. We therefore use two-sided Fisher exact tests for $AR$ and $rAR$ product comparisons in this subset, while $PR$ is $1.0$ by construction.
Pairwise product comparisons are corrected using Holm-Bonferroni correction within each image set and metric.
To assess annotation reliability, we compute a generalized Fleiss' $\kappa$ from the clean, artifact, and text-only artifact vote counts.
Finally, we report version-level confidence intervals to summarize whether repeated regeneration attempts reduce artifact prevalence. These version-level rates are interpreted descriptively because later versions are generated conditional on earlier judge failures.
\section{Results}
\label{sec:results}
Below, we present the results of the qualitative and quantitative evaluation. In \Cref{subsec:qualitative-results}, we report the qualitative feedback of the focus group, which was conducted with outputs from the OpenAI GPT configuration and includes the named strengths and challenges of generated images. We continue with the quantitative results in \Cref{subsec:quantitative-results}, where we report $PR$, $AR$, and $rAR$ for all subsets, products, and model configurations.

\subsection{Qualitative results}
\label{subsec:qualitative-results}

The focus group identified a recurring set of strengths and challenges across the generated advertisements from the OpenAI GPT configuration. Overall, this qualitative evaluation supports the feasibility of personalization: experts observed that persona attributes such as hobbies, pets, cultural background, appearance, and lifestyle cues were reflected in the generated scenes. \Cref{fig:image_grid_celebs} shows representative outputs for the six celebrity personas across all four products. The following paragraphs summarize the aggregated observations.

\newlength{\gridwidth}
\newcommand{\gridimage}[1]{\includegraphics[width=\linewidth]{#1}}

\begin{figure*}[tbp]
\centering
\setlength{\gridwidth}{0.8\linewidth}   
\scriptsize
\setlength{\tabcolsep}{1pt}
\renewcommand{\arraystretch}{1.0}
\renewcommand{\tabularxcolumn}[1]{m{#1}}

\begin{tabularx}{\gridwidth}{@{}>{\centering\arraybackslash}m{0.035\gridwidth} *{6}{>{\centering\arraybackslash}X}@{}}
& \shortstack{Bruce\\Wayne\\(The Batman)}
& \shortstack{Morpheus\\(The\\Matrix)}
& \shortstack{Vito\\Corleone\\(The\\Godfather)}
& \shortstack{Yan Naing\\Lee\\(Rush Hour)}
& \shortstack{Rachel\\Green\\(Friends)}
& \shortstack{Rajesh\\(Big Bang\\Theory)} \\

\rotatebox[origin=c]{90}{Electric vehicle} &
\gridimage{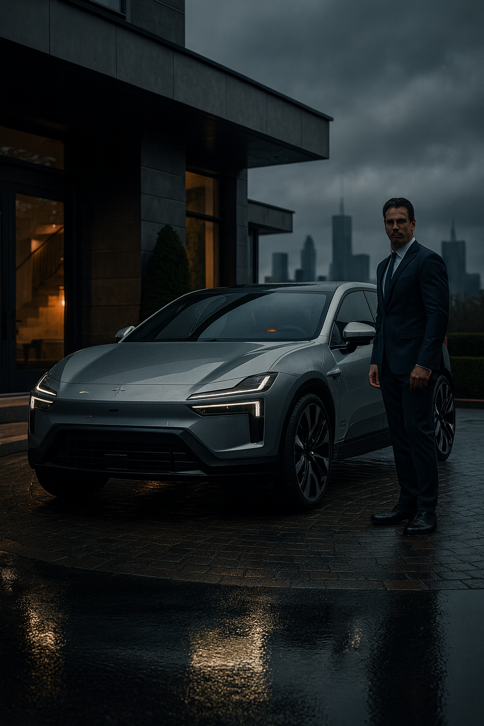} &
\gridimage{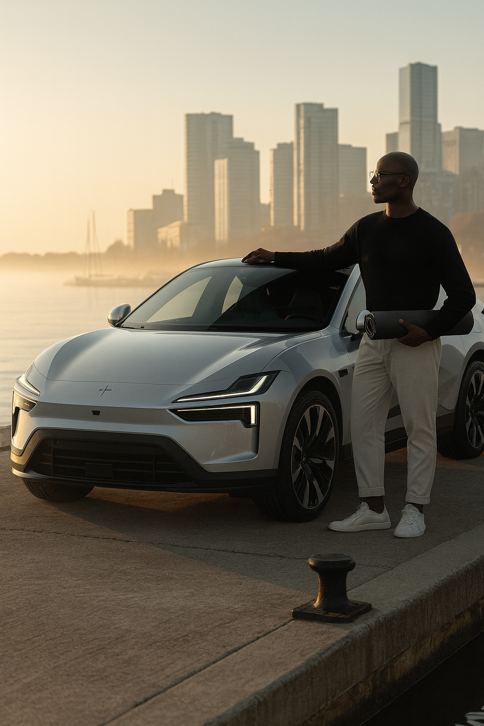} &
\gridimage{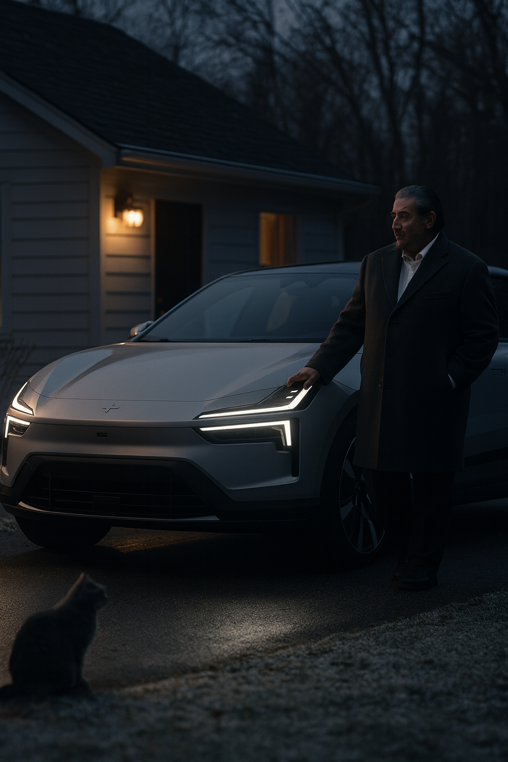} &
\gridimage{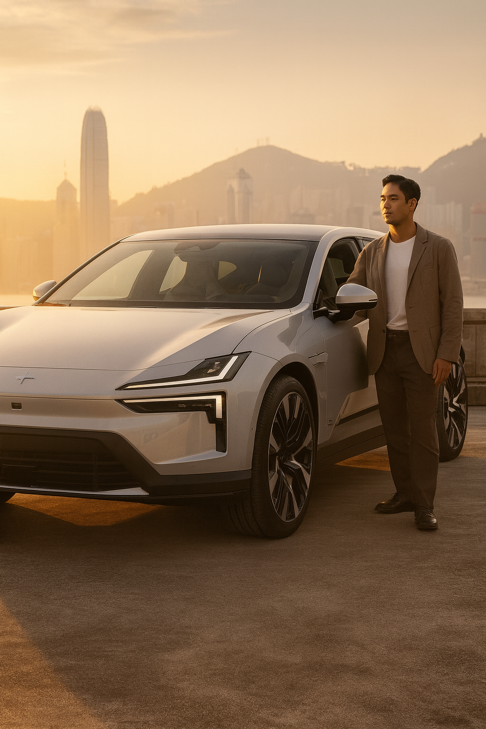} &
\gridimage{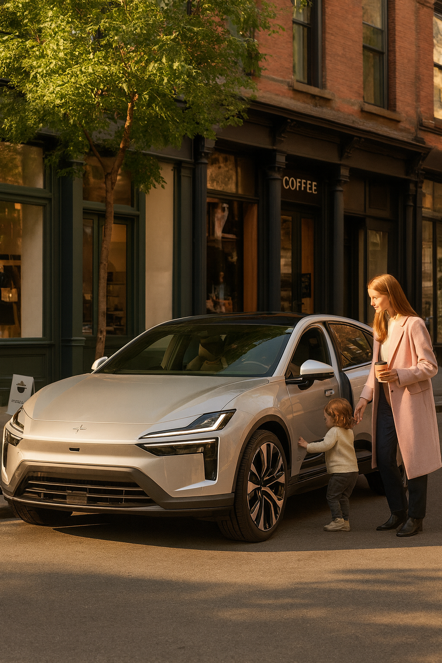} &
\gridimage{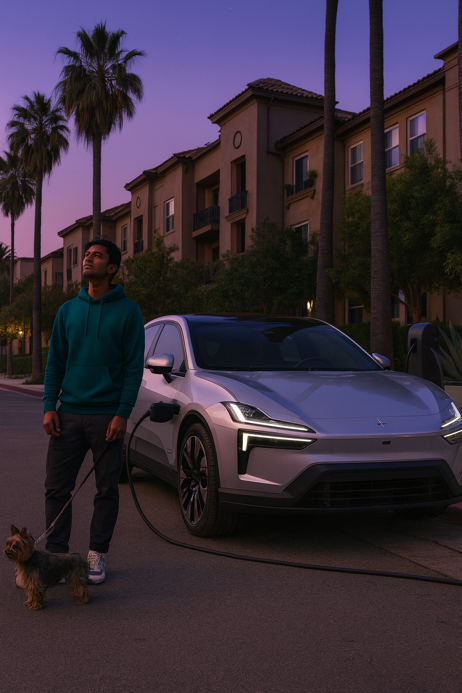} \\
\noalign{\vskip 1pt}

\rotatebox[origin=c]{90}{Luxury watch} &
\gridimage{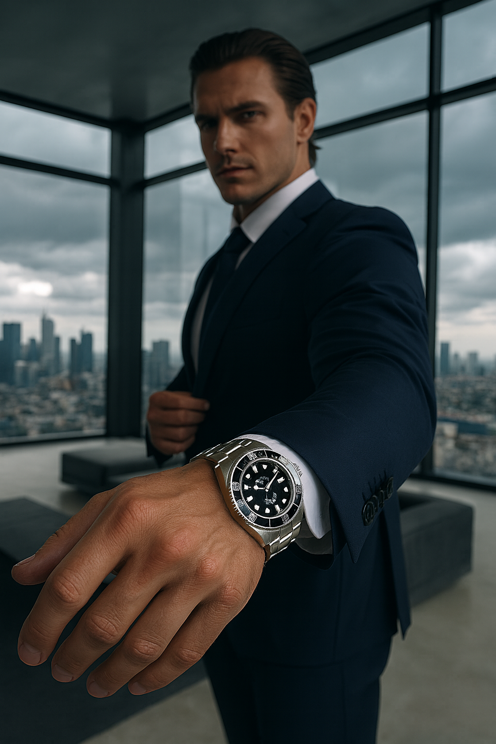} &
\gridimage{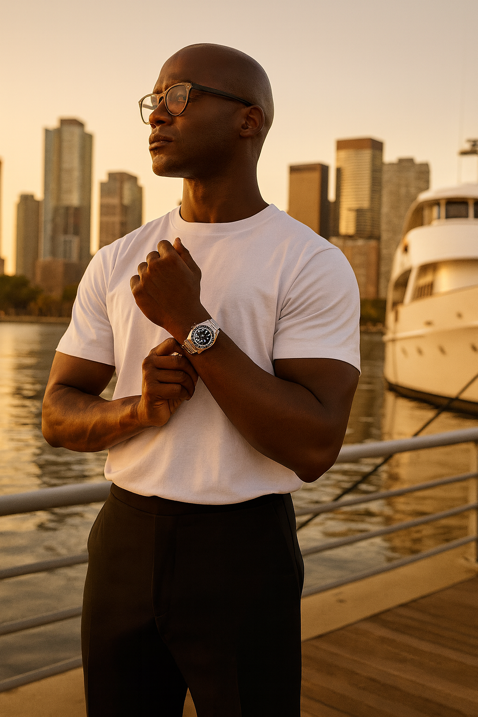} &
\gridimage{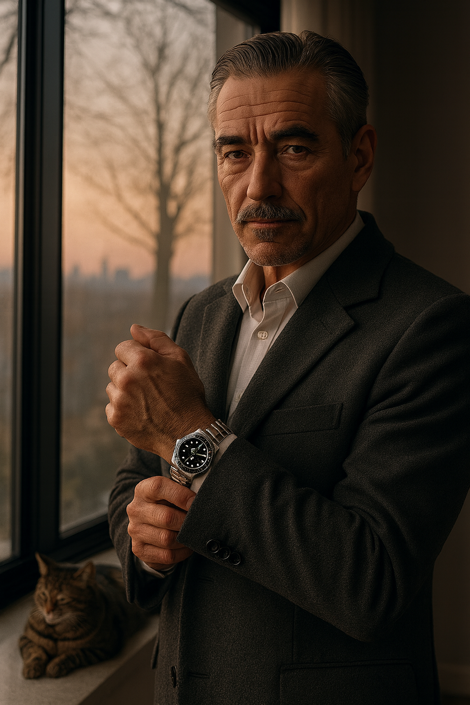} &
\gridimage{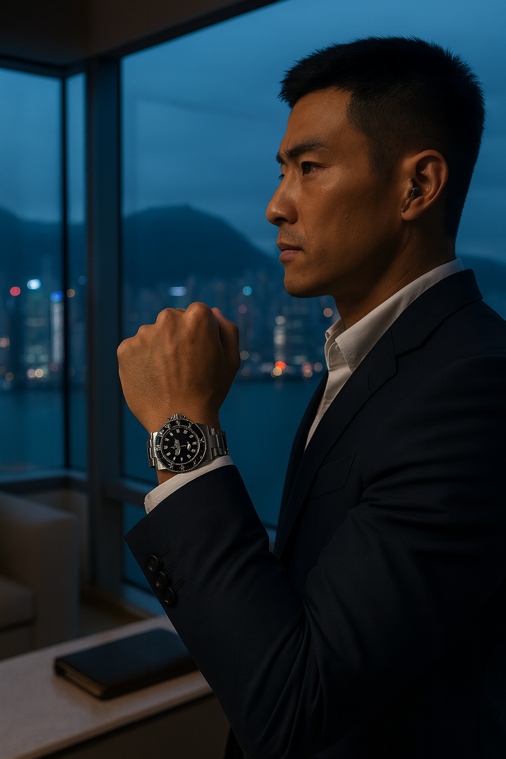} &
\gridimage{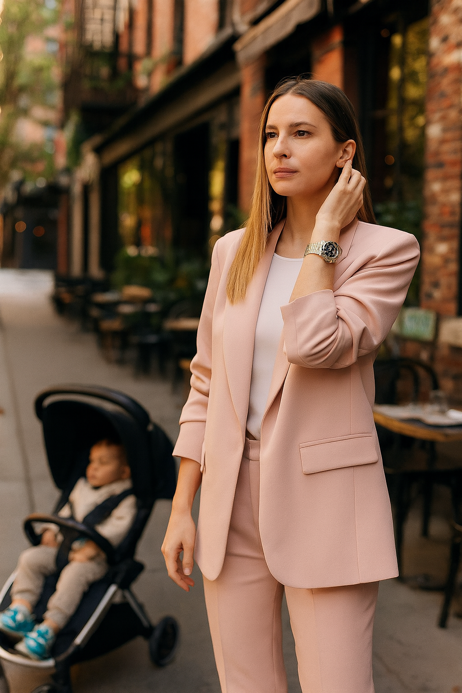} &
\gridimage{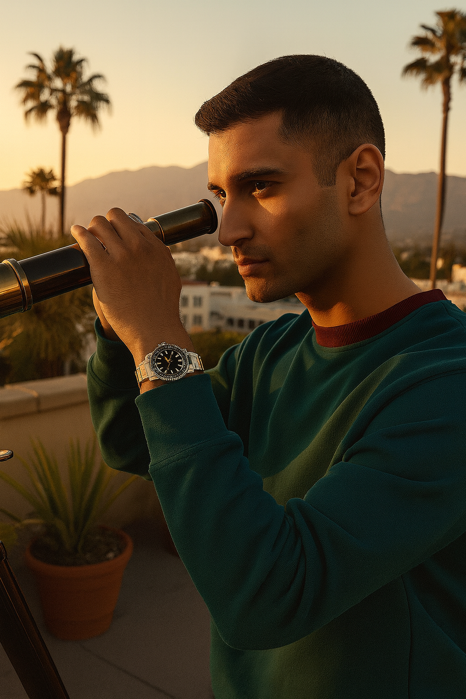} \\
\noalign{\vskip 1pt}

\rotatebox[origin=c]{90}{Laundry detergent} &
\gridimage{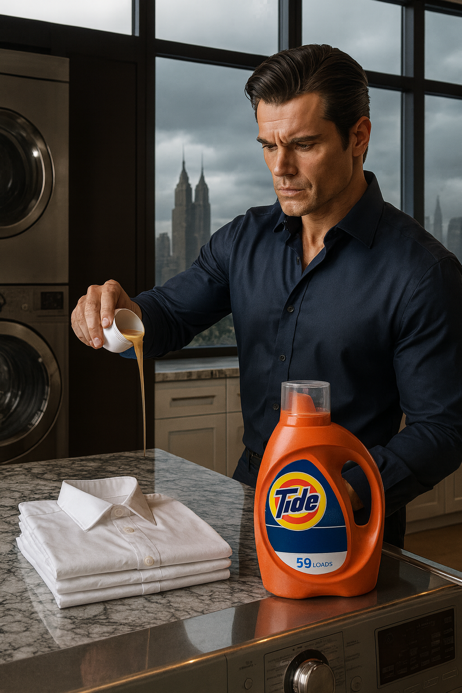} &
\gridimage{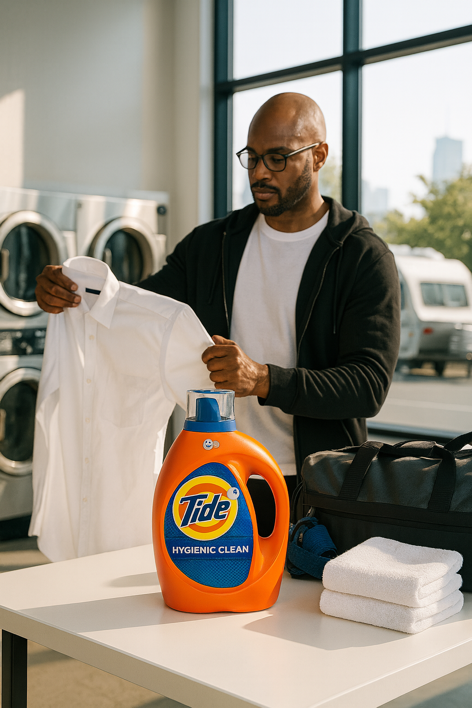} &
\gridimage{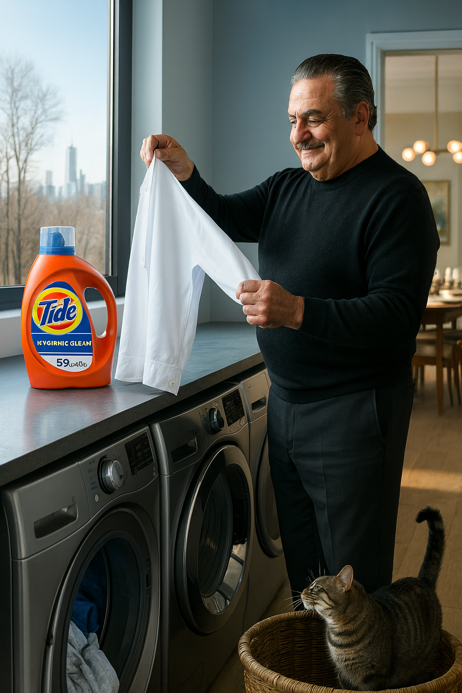} &
\gridimage{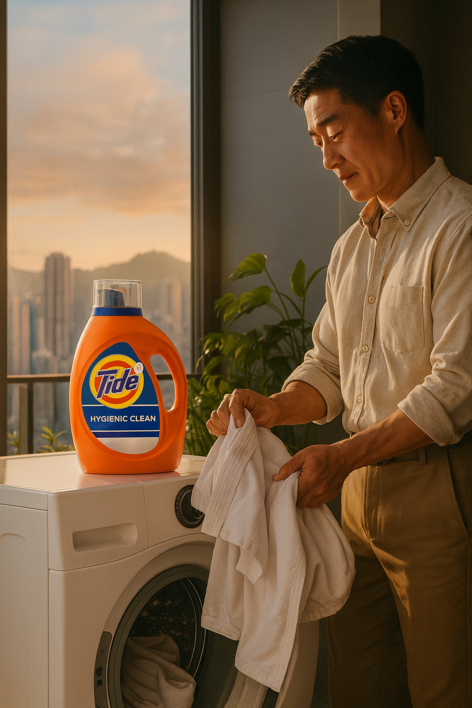} &
\gridimage{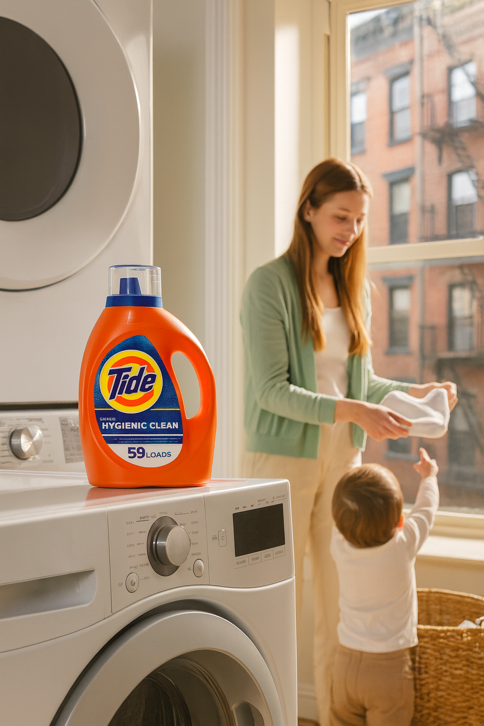} &
\gridimage{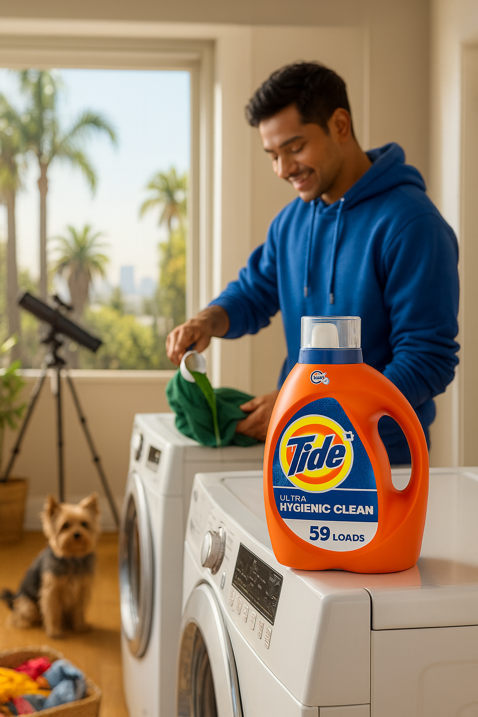} \\
\noalign{\vskip 1pt}

\rotatebox[origin=c]{90}{Soft drink} &
\gridimage{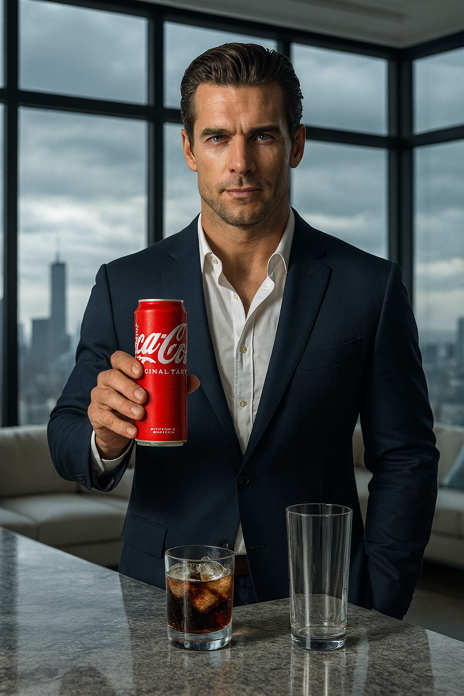} &
\gridimage{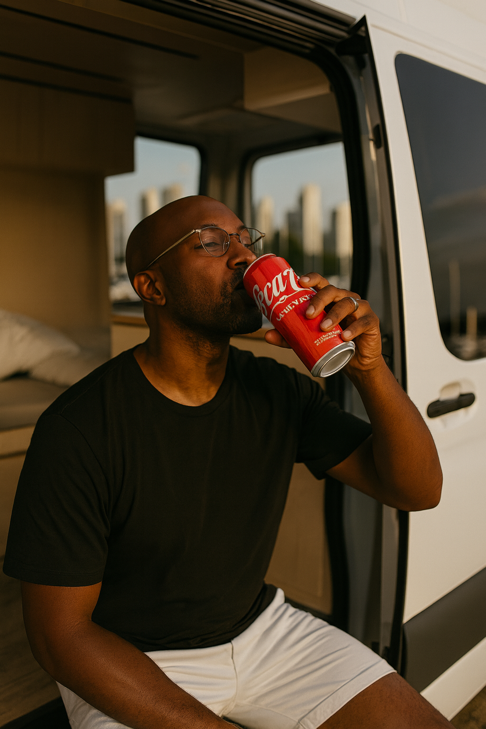} &
\gridimage{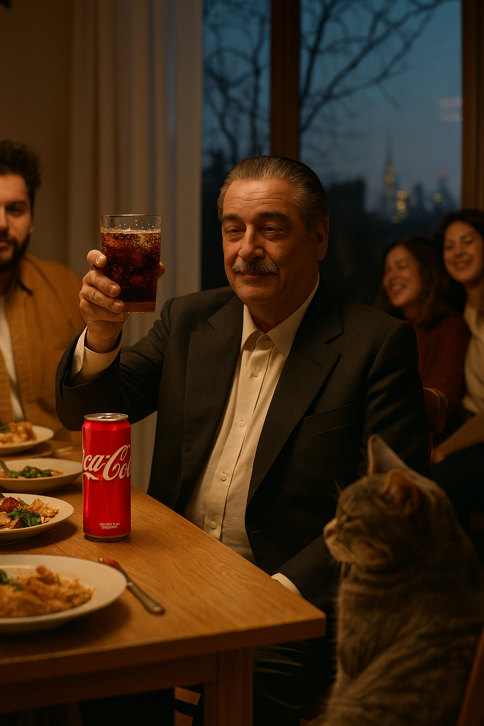} &
\gridimage{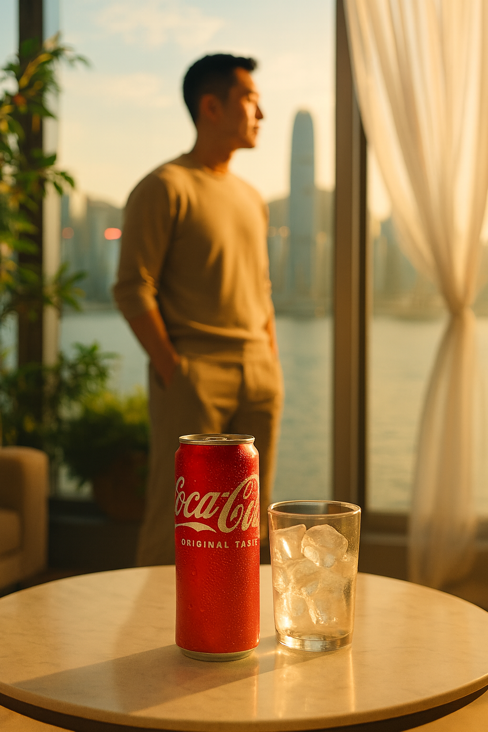} &
\gridimage{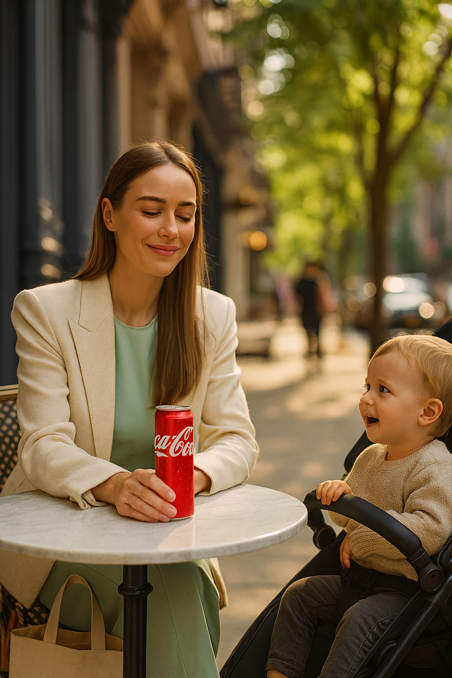} &
\gridimage{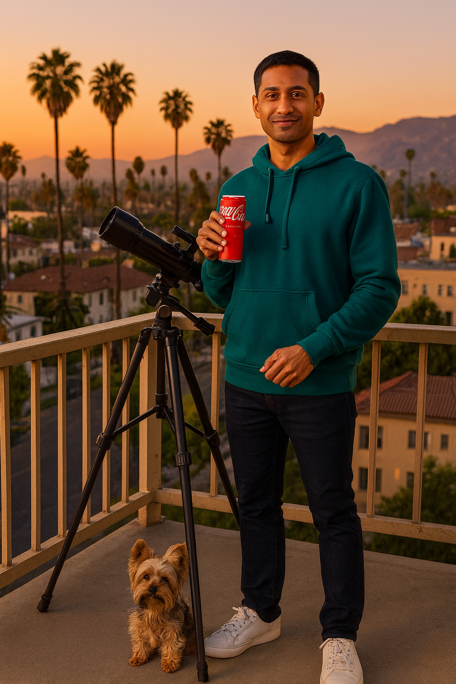} \\
\end{tabularx}

\caption{Advertisements from the qualitative evaluation for all six characters (columns) generated for the four products (rows) using the OpenAI GPT model configuration.}
\label{fig:image_grid_celebs}
\end{figure*}

\paragraph{Strengths.}

Experts consistently note the high \textbf{visual quality and photographic realism} of the generated images. Multiple experts highlighted convincing depth-of-field effects, well-chosen camera angles, natural lighting, and fine-grained rendering of surfaces such as fabric, reflections, and wrinkles. These properties contributed to an overall professional aesthetic that was considered comparable to photography in several cases.
A second recurring strength was \textbf{scene coherence and narrative plausibility}. Several images were praised for conveying a convincing lifestyle story aligned with the persona---for instance, a family reference in an electric vehicle scene and a cat or dog appearing in images whose persona descriptions included pet ownership. Experts noted that the camera perspective and scene composition often placed the viewer \enquote{right in the action}, contributing to an immersive, authentic feel.
A third strength concerns \textbf{contextual and cultural accuracy}. In at least one instance, a city-specific landmark (a well-known skyscraper) was correctly incorporated, demonstrating that cultural signals from persona descriptions can propagate meaningfully into the visual output. Appropriate clothing and setting choices (e.g., suited clothing for a high-income persona in a private environment) further supported this observation.
Finally, experts noted that \textbf{several individual images} across all four product categories \textbf{were rated as visually compelling} and suitable as advertisements. The Coca-Cola, Rolex, and Polestar outputs in particular received positive overall assessments, and the Tide laundry detergent image was noted for depicting underrepresented characteristics (e.g., a heavier-set, stay-at-home male figure) in a non-stereotypical manner.

\paragraph{Challenges.}

The most frequently raised challenge concerned \textbf{structural artifacts and object-level defects}. Experts identified missing or incorrectly rendered structural elements, including absent washing machine doors, a car door appearing simultaneously open and closed, a misplaced vehicle charging port, and a figure appearing partially inside a washing machine. Towels were observed to fall from a door that appeared closed, and shirt folding was noted as visually glitched in at least one image. These defects directly reduce the usability of the generated image as advertisements.
A second challenge relates to \textbf{unnatural anatomy and body postures}. Experts noted awkward standing positions near vehicles, excessively wide hands or fingers, unnatural eye appearance, and implausible arm or hand poses when holding objects (e.g., a telescope or a luxury watch). While isolated cases were rated positively (e.g., one image was noted for natural, athletic body language), unnatural postures represented a recurring failure mode across multiple persona--product combinations.
Third, experts identified issues with \textbf{product rendering and placement}. The laundry detergent bottle was described as appearing artificial or placed as if on a green screen. The Coca-Cola can was rated as disproportionately large, and its label contained incorrect lettering. The luxury watch appeared unnatural when depicted on a person's arm.
A fourth challenge concerns \textbf{scene plausibility and authenticity}. Several compositions were characterized as overly constructed or artificial. A stroller was noted to be placed implausibly in an empty space, and a figure was described as abandoning the stroller to pose with a watch in a way that was considered inauthentic. Some scenes featuring social settings did not reflect the social activities mentioned in the persona description, with one image depicting a figure resembling a single parent rather than a person engaged in socializing.
Fifth, experts observed \textbf{demographic and age inconsistencies} between generated figures and persona descriptions. In some cases, the depicted person appeared younger or older than the age specified in the persona. Income level was not consistently reflected in the visual representation of settings or objects.
Sixth, several experts noted \textbf{repetition and compositional anomalies}, including multiple washing machines appearing within a single image and stacked appliances producing a visually implausible arrangement.
Finally, experts flagged \textbf{biases} in the generated outputs. Gender- and age-stereotypical representations were observed in some images (e.g., a luxury watch image reproducing stereotypes of an \enquote{attractive best ager}), although experts acknowledged that some of these outcomes may be attributable to the input persona descriptions rather than the model itself. Across the set of generated images, children were noted to be overrepresented relative to their prominence in persona descriptions.

\subsection{Quantitative results}
\label{subsec:quantitative-results}

\Cref{tab:quantitative_results} reports the $PR$ from the judge agent, the $AR$, and the $rAR$ across the three image sets $I_{\text{first}}$, $I_{\text{passed}}$, and $I_{\text{selected}}$ for both model configurations. The $AR$ and $rAR$ provide a human-labeled assessment of generated image quality, whereas $PR$ captures the judge agent's pass/fail decision. For the OpenAI GPT configuration, the first generation attempt $I_{\text{first}}$ achieved an average $AR=0.50$, indicating that half of the first generated images contained artifacts. The lower $rAR=0.27$ indicates that a substantial share of artifacts is attributable to text rendering. The corresponding initial pass rate is $PR=0.51$, indicating that approximately half of the initial images were accepted without regeneration. The judge-filtered set $I_{\text{passed}}$ exhibits a lower artifact prevalence ($AR=0.35$, $rAR=0.19$). Because membership in this set is conditional on the judge's approval, the difference reflects how well the judge sorts the available candidates rather than an improvement produced by regeneration itself; we therefore test the judge's discrimination directly below rather than comparing the image sets against each other. The final selected set $I_{\text{selected}}$ has $PR=0.79$, $AR=0.42$, and $rAR=0.21$, making it worse than $I_{\text{passed}}$ on average because the regeneration process stops after three attempts and may therefore deliver a best-of candidate even if no version passes the judge.

The Google Gemini configuration shows a different pattern. Its first generated images have substantially lower human-labeled artifact rates than the OpenAI GPT configuration ($AR=0.23$, $rAR=0.05$ in $I_{\text{first}}$), suggesting stronger image-generation performance with respect to visible artifacts, especially non-text artifacts. After judge filtering, artifact prevalence decreases further ($AR=0.04$, $rAR=0.04$ in $I_{\text{passed}}$). The final selected set reaches $AR=0.20$ and $rAR=0.04$, which remains lower than the OpenAI GPT configuration. At the same time, Gemini's pass rates are lower ($PR=0.22$ in $I_{\text{first}}$ and $PR=0.40$ in $I_{\text{selected}}$), indicating that the Gemini judge rejects a larger share of images despite their lower human-labeled artifact prevalence. Thus, the table points to two distinct effects: the Gemini image generator produces fewer artifact-labeled images on average, while the Gemini judge applies stricter pass criteria within the regeneration limit.

\begin{table*}[tbh]

\centering
\footnotesize
\setlength{\tabcolsep}{2pt}
\renewcommand{\arraystretch}{1.05}

\begin{tabular*}{\textwidth}{@{\extracolsep{\fill}} l c c c c c c c c c @{}}
\toprule
& \multicolumn{3}{c}{\shortstack{\textbf{First generation}\\($I_{\text{first}}$)}} & \multicolumn{3}{c}{\shortstack{\textbf{Judge passed}\\($I_{\text{passed}}$)}} & \multicolumn{3}{c}{\shortstack{\textbf{Final selected}\\($I_{\text{selected}}$)}} \\
\cmidrule(lr){2-4}\cmidrule(lr){5-7}\cmidrule(lr){8-10}
\textbf{Product} & $\mathbf{PR}$ & $\mathbf{AR}$ & $\mathbf{rAR}$ & $\mathbf{PR}$ & $\mathbf{AR}$ & $\mathbf{rAR}$ & $\mathbf{PR}$ & $\mathbf{AR}$ & $\mathbf{rAR}$ \\
\midrule
\multicolumn{10}{@{}l}{\textbf{OpenAI GPT configuration}} \\
Electric Vehicle (P1)     & 0.86 & 0.40 & 0.40 & 1.00 & 0.37 & 0.37 & 1.00 & 0.37 & 0.37 \\
Luxury Watch (P2)         & 0.57 & 0.20 & 0.20 & 1.00 & 0.07 & 0.07 & 0.88 & 0.08 & 0.08 \\
Laundry Detergent (P3)    & 0.20 & 0.84 & 0.45 & 1.00 & 0.60 & 0.34 & 0.50 & 0.69 & 0.36 \\
Soft Drink (P4)           & 0.41 & 0.56 & 0.01 & 1.00 & 0.49 & 0.01 & 0.76 & 0.56 & 0.03 \\
\midrule
\textbf{Overall Mean} & \textbf{0.51} & \textbf{0.50} & \textbf{0.27} & \textbf{1.00} & \textbf{0.35} & \textbf{0.19} & \textbf{0.79} & \textbf{0.42} & \textbf{0.21} \\
\midrule
\multicolumn{10}{@{}l}{\textbf{Google Gemini configuration}} \\
Electric Vehicle (P1)     & 0.56 & 0.03 & 0.03 & 1.00 & 0.04 & 0.04 & 0.89 & 0.05 & 0.05 \\
Luxury Watch (P2)         & 0.18 & 0.04 & 0.04 & 1.00 & 0.06 & 0.06 & 0.34 & 0.04 & 0.04 \\
Laundry Detergent (P3)    & 0.01 & 0.81 & 0.11 & 1.00 & 0.33 & 0.00 & 0.03 & 0.69 & 0.07 \\
Soft Drink (P4)           & 0.14 & 0.05 & 0.03 & 1.00 & 0.00 & 0.00 & 0.32 & 0.03 & 0.02 \\
\midrule
\textbf{Overall Mean} & \textbf{0.22} & \textbf{0.23} & \textbf{0.05} & \textbf{1.00} & \textbf{0.04} & \textbf{0.04} & \textbf{0.40} & \textbf{0.20} & \textbf{0.04} \\
\bottomrule
\end{tabular*}
\caption{Quantitative results reporting pass rate ($PR$), artifact rate ($AR$), and relaxed artifact rate ($rAR$) for the OpenAI GPT and Google Gemini configurations across three image sets: the initial attempt ($I_{\text{first}}$), the subset of images approved by the judge ($I_{\text{passed}}$), and the final image delivered to the customer ($I_{\text{selected}}$).}
\label{tab:quantitative_results}
\end{table*}

Across products, image-generation performance varies substantially in both configurations. In the OpenAI GPT configuration, the electric vehicle (P1) has the highest initial pass rate ($PR=0.86$) and all final selected images pass ($PR=1.00$), while the AR decreases only modestly from $AR=0.40$ in $I_{\text{first}}$ to $0.37$ in $I_{\text{selected}}$. The luxury watch (P2) shows a stronger reduction in artifacts after judge filtering ($AR=0.20$ in $I_{\text{first}}$ vs. $0.07$ in $I_{\text{passed}}$). In contrast, laundry detergent (P3) is the most difficult product for the OpenAI GPT image generator, with the highest initial artifact prevalence ($AR=0.84$), a relatively high non-text artifact prevalence ($rAR=0.45$), and a final selected set that still contains many artifacts ($AR=0.69$, $rAR=0.36$). For the soft drink (P4), the large gap between $AR$ and $rAR$ indicates that many defects are text-related.

For the Google Gemini configuration, the image generator achieves low artifact rates for P1, P2, and P4 across image sets, and the pairwise tests reported below show that these three products are statistically indistinguishable from one another. Their final artifact rates are $AR=0.05$ for the electric vehicle, $AR=0.04$ for the luxury watch, and $AR=0.03$ for the soft drink, which also has the lowest relaxed rate ($rAR=0.02$ in $I_{\text{selected}}$). A plausible reading is that Gemini generates comparatively clean images for products with simpler visual structures or less complex object interactions, although product complexity is not manipulated experimentally and this interpretation therefore remains descriptive. Laundry detergent (P3), however, again emerges as the most challenging product: the final selected set retains a high artifact rate ($AR=0.69$), even though the relaxed artifact rate is much lower ($rAR=0.07$). This suggests that, across both image generators, laundry detergent is the product category in which repeated generation most often fails to produce a fully acceptable final image. The near-zero initial pass rate for Gemini on this product ($PR=0.01$) further indicates that the judge is particularly strict for this product category.

A key pattern across products is that artifacts are frequently driven by text rendering on products and packaging. This pattern is strongest in the OpenAI GPT configuration, where large gaps between $AR$ and $rAR$ appear for laundry detergent (P3) and soft drink (P4). For P3, $I_{\text{first}}$ yields $AR=0.84$ but $rAR=0.45$, and $I_{\text{selected}}$ yields $AR=0.69$ but $rAR=0.36$. For P4, the gap is even more pronounced: $I_{\text{first}}$ yields $AR=0.56$ but $rAR=0.01$, and $I_{\text{selected}}$ yields $AR=0.56$ but $rAR=0.03$. In the Google Gemini configuration, large $AR$--$rAR$ gaps are concentrated primarily in laundry detergent: $I_{\text{first}}$ yields $AR=0.81$ but $rAR=0.11$, and $I_{\text{selected}}$ yields $AR=0.69$ but $rAR=0.07$. This indicates that incorrect or unrealistic text remains a dominant failure mode, although its prevalence differs substantially by model configuration and product category. The two cells with the widest gaps, soft drink under OpenAI GPT and laundry detergent under Google Gemini, are also the two in which the reviewers agreed least, as the inter-rater reliability results below show. The direction of these gaps is therefore well supported, but their exact magnitude carries more annotation uncertainty than the remaining cells.

The final selected pass rates imply different operational trade-offs. For OpenAI GPT, the overall mean pass rate of $PR=0.79$ in $I_{\text{selected}}$ implies that in $21\%$ of customer-product pairs, no generated version passed within the version limit and the pipeline delivered a best-of candidate despite failing the judge's criteria. For Google Gemini, the corresponding final selected pass rate is lower ($PR=0.40$), implying that $60\%$ of final deliveries are best-of selections rather than judge-approved outputs. The set $I_{\text{selected}}$ therefore captures not only artifact detection, but also the judge's second task: selecting the best available image when all candidates fail. However, the final artifact rates for Gemini remain lower on average ($AR=0.20$, $rAR=0.04$) than for OpenAI GPT ($AR=0.42$, $rAR=0.21$). This suggests that pass decisions and artifact labels capture related but distinct aspects of output quality: the image generation model determines the artifact profile of the candidates, while the judge determines which candidates are accepted, rejected, or selected as the best available fallback. This behavior is illustrated by the image examples and judge rationales (\Cref{fig:example_images_bad,fig:judge_reasoning_examples}) as well as by the version comparison under regeneration (\Cref{fig:version_comparison_grid_final_spanning}), where repeated candidates can fail primarily due to persistent packaging or text errors. In such cases, best-of selection can choose the comparatively strongest candidate, but it cannot compensate for a candidate pool in which all generated versions contain commercially relevant artifacts. In addition, the judge agent is unable to detect some artifacts because it lacks a real physical understanding, as illustrated in~\Cref{fig:judge_reasoning_examples} by the obvious artifact showing the electric vehicle's reversed trunk.

\paragraph{Statistical validation}
\Cref{tab:statistical_validation} reports the overall rates of \Cref{tab:quantitative_results} with Wilson $95\%$ confidence intervals and the paired comparison of the two model configurations. Except for $PR$ in $I_{\text{passed}}$, which equals $1.0$ by construction, the intervals of the two configurations are disjoint for every metric and image set, and all six paired comparisons are significant in exact McNemar tests ($p<0.001$). The Google Gemini configuration produces fewer human-labeled artifacts ($\Delta AR = -0.22$, $95\%$ CI $[-0.28$, $-0.17]$ in $I_{\text{selected}}$) but passes markedly fewer images ($\Delta PR = -0.39$, $[-0.44$, $-0.34]$). The two effects described above---a cleaner image generator and a stricter judge in the Google Gemini configuration---are therefore statistically robust rather than an artifact of the descriptive means. The same separation holds in $I_{\text{passed}}$, which is excluded from \Cref{tab:statistical_validation} because its membership depends on the judge's decisions and its denominators therefore differ ($n=314$ vs. $n=158$): the artifact rate is $0.35$, $[0.30$, $0.41]$ for the OpenAI GPT configuration and $0.04$, $[0.02$, $0.09]$ for the Google Gemini configuration. Because each customer contributes four image pairs, we repeat the comparison on customer-level mean differences; the clustered intervals and the exact sign tests across the $100$ customers lead to the same conclusions ($p<0.001$ for all six comparisons).

\begin{table*}[tb]

\centering
\footnotesize
\setlength{\tabcolsep}{3pt}
\renewcommand{\arraystretch}{1.05}

\begin{tabular*}{\textwidth}{@{\extracolsep{\fill}} l l c c c c @{}}
\toprule
\textbf{Image set} & \textbf{Metric} & \shortstack{\textbf{OpenAI GPT}\\\textbf{[95\% CI]}} & \shortstack{\textbf{Google Gemini}\\\textbf{[95\% CI]}} & $\mathbf{\Delta}$ & $\mathbf{p}$ \\
\midrule
$I_{\text{first}}$ & $PR$  & 0.51 [0.46, 0.56] & 0.22 [0.18, 0.27] & $-0.29$ & $<0.001$ \\
$I_{\text{first}}$ & $AR$  & 0.50 [0.45, 0.55] & 0.23 [0.19, 0.28] & $-0.27$ & $<0.001$ \\
$I_{\text{first}}$ & $rAR$ & 0.27 [0.22, 0.31] & 0.05 [0.03, 0.08] & $-0.21$ & $<0.001$ \\
$I_{\text{selected}}$ & $PR$  & 0.79 [0.74, 0.82] & 0.40 [0.35, 0.44] & $-0.39$ & $<0.001$ \\
$I_{\text{selected}}$ & $AR$  & 0.42 [0.38, 0.47] & 0.20 [0.17, 0.24] & $-0.22$ & $<0.001$ \\
$I_{\text{selected}}$ & $rAR$ & 0.21 [0.17, 0.25] & 0.04 [0.03, 0.07] & $-0.17$ & $<0.001$ \\
\bottomrule
\end{tabular*}
\caption{Statistical validation of the overall rates for the paired image sets. Wilson $95\%$ confidence intervals are reported in brackets. $\Delta$ is the paired difference (Google Gemini minus OpenAI GPT), and $p$ is from two-sided exact McNemar tests at the customer--product level ($n=400$ pairs per row). Negative differences indicate lower rates for Google Gemini, which is favorable for $AR$ and $rAR$ but unfavorable for $PR$. Because $\Delta$ is computed from unrounded rates, it can deviate by $0.01$ from the difference of the rounded rates shown. $I_{\text{passed}}$ is not included because its membership depends on the judge's decisions and it is therefore not paired across configurations; its rates are reported in \Cref{tab:quantitative_results} and its intervals in the text.}
\label{tab:statistical_validation}
\end{table*}

The pairwise product comparisons reveal a systematic difference in how the two configurations fail. For the OpenAI GPT configuration, product differences are pervasive: after Holm correction, at least four of six contrasts remain significant for every metric and image set in which the metric varies. For the Google Gemini configuration, pass rates differ across products just as systematically, but the differences in artifact prevalence are concentrated almost entirely on a single product. In both $I_{\text{first}}$ and $I_{\text{selected}}$, exactly the three $AR$ contrasts that involve laundry detergent are significant (all $p_{\text{Holm}}<0.001$), whereas none of the three contrasts among electric vehicle, luxury watch, and soft drink is, and not a single $rAR$ contrast survives correction in any image set. Statistically, the Google Gemini configuration therefore has one hard product rather than a graded ordering of product difficulty. Its $I_{\text{passed}}$ values for laundry detergent rest on only three images ($AR=0.33$, $[0.06$, $0.79]$) and should not be over-interpreted.

The judge's pass decisions align with the human artifact labels, but almost entirely through text. Within $I_{\text{first}}$, where judge-passed and judge-failed images come from the same $400$ customer--product pairs and are separated only by the judge's decision, passed images have a substantially lower $AR$ than failed images in both configurations ($0.32$ vs. $0.69$ for OpenAI GPT and $0.03$ vs. $0.29$ for Google Gemini, both $p<0.001$). For $rAR$, which excludes text-only defects, the separation is much weaker for the OpenAI GPT judge ($0.21$ vs. $0.33$, $p=0.007$) and not detectable for the Google Gemini judge ($0.03$ vs. $0.06$, $p=0.589$), although the small number of images that the Gemini judge passes ($n=89$) limits the power of this particular test. Both judges therefore act largely as detectors of text defects, and neither shows more than weak evidence of detecting structural artifacts. This quantifies the qualitative observation that the judge misses obvious physical implausibilities (\Cref{fig:judge_reasoning_examples}), and it explains why the stricter Google Gemini judge does not translate into cleaner final deliveries with respect to non-text defects.

Annotation reliability is moderate to substantial and stable across image sets. The generalized Fleiss' $\kappa$ is $0.548$ for the OpenAI GPT configuration and $0.647$ for the Google Gemini configuration. It is lowest in the two product--configuration cells with the largest gaps between $AR$ and $rAR$, namely soft drink under OpenAI GPT ($\kappa=0.171$) and laundry detergent under Google Gemini ($\kappa=0.318$), where reviewers disagree almost exclusively about whether rendered text is defective rather than about structural defects. This supports reading $rAR$ as the more reliable of the two artifact measures.

Finally, repeated generation does not reduce artifact prevalence within the attempt limit. Across the three versions, $AR$ rises from $0.50$ to $0.56$ and $0.73$ for the OpenAI GPT configuration and from $0.23$ to $0.28$ and $0.29$ for the Google Gemini configuration, and in neither configuration does a later version reach an artifact rate whose confidence interval lies below that of the first version. As stated in \Cref{subsubsec:artifact_rate}, these rates are descriptive, because later versions exist only for customer--product pairs whose preceding version was rejected by the judge and the samples are therefore progressively selected toward the harder pairs.

\begin{figure*}[tbp]
\centering
\begingroup
\scriptsize
\setlength{\tabcolsep}{2pt}
\renewcommand{\arraystretch}{1.0}
\newcommand{\exampleimage}[1]{\begin{minipage}[t]{\linewidth}\vspace{0pt}\centering\includegraphics[width=\linewidth]{#1}\end{minipage}}
\newcommand{\examplejudgement}[1]{\begin{minipage}[t]{\linewidth}\vspace{0pt}\raggedright #1\end{minipage}}

\begin{tabular}{@{}>{\centering\arraybackslash}p{0.09\linewidth} p{0.375\linewidth} >{\centering\arraybackslash}p{0.09\linewidth} p{0.375\linewidth}@{}}
\toprule
\multicolumn{2}{@{}c}{\textbf{OpenAI GPT configuration}} &
\multicolumn{2}{c@{}}{\textbf{Google Gemini configuration}} \\
\cmidrule(lr){1-2}\cmidrule(lr){3-4}
\textbf{Image} & \textbf{Judge decision \& reasoning} &
\textbf{Image} & \textbf{Judge decision \& reasoning} \\
\midrule

\exampleimage{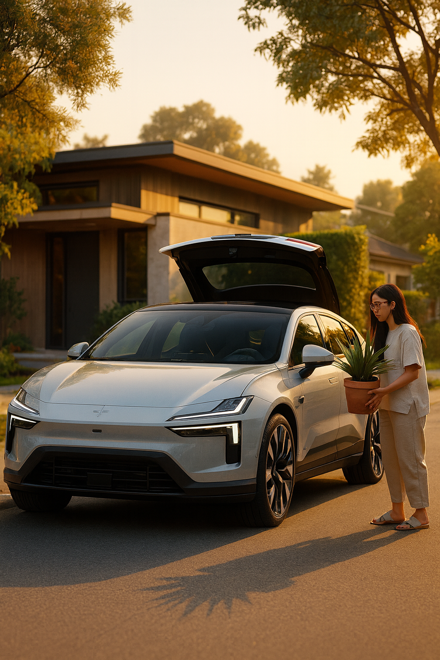} &
\examplejudgement{\textbf{PASS.} ``High-quality, photo-realistic lifestyle auto image at golden hour. Vehicle geometry, human anatomy, and environment are largely consistent and believable. Only minor issues involving lighting/shadow consistency and subtle tailgate detailing prevent a perfect score.''} &
\exampleimage{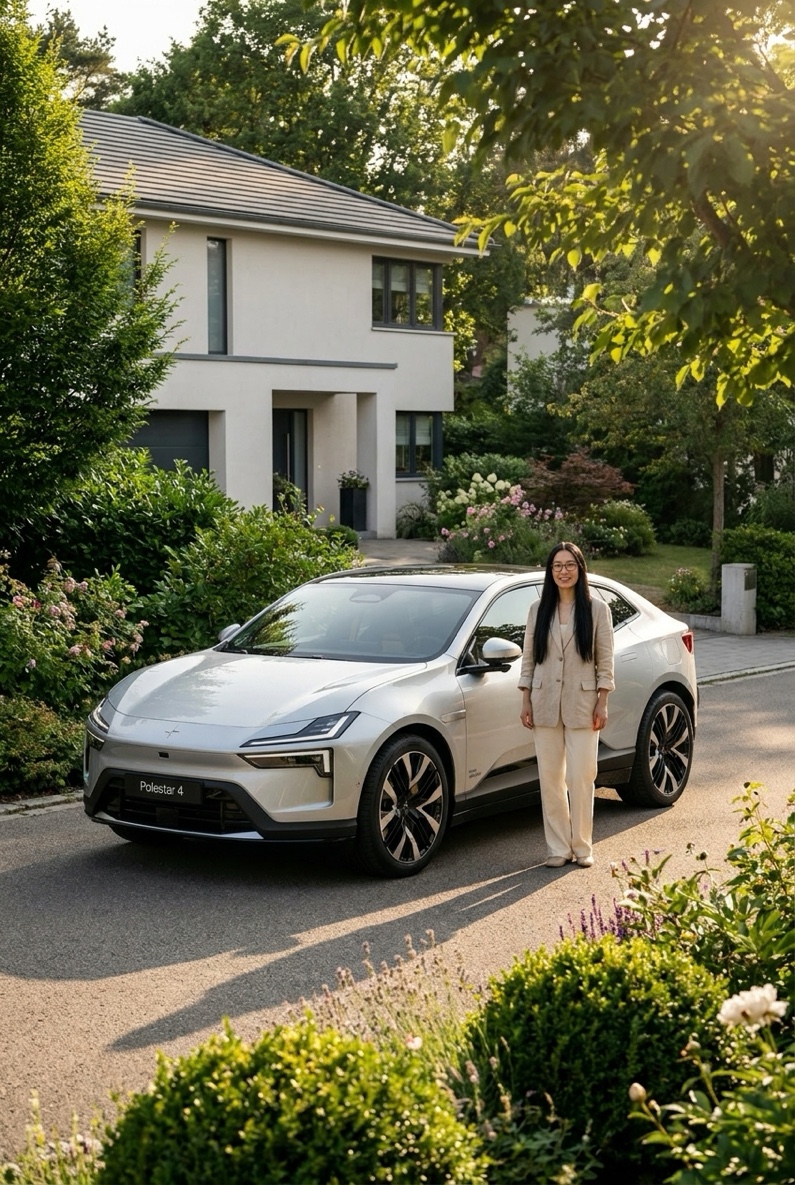} &
\examplejudgement{\textbf{PASS.} ``The image demonstrates high-quality generation with no major anatomical or physical anomalies. The structural integrity of the car, person, and background elements is well-maintained.''} \\

\exampleimage{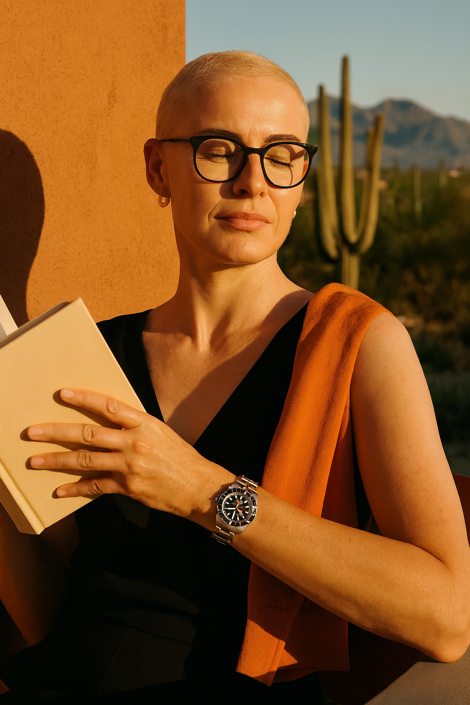} &
\examplejudgement{\textbf{PASS.} ``High-quality, realistic lifestyle shot with consistent golden-hour lighting, coherent shadows, and a shallow depth of field. The watch is prominent and integrated believably. Only two minor issues: slight bracelet geometry/fit artifact and a seconds-hand color that doesn't match the clean product shot.''} &
\exampleimage{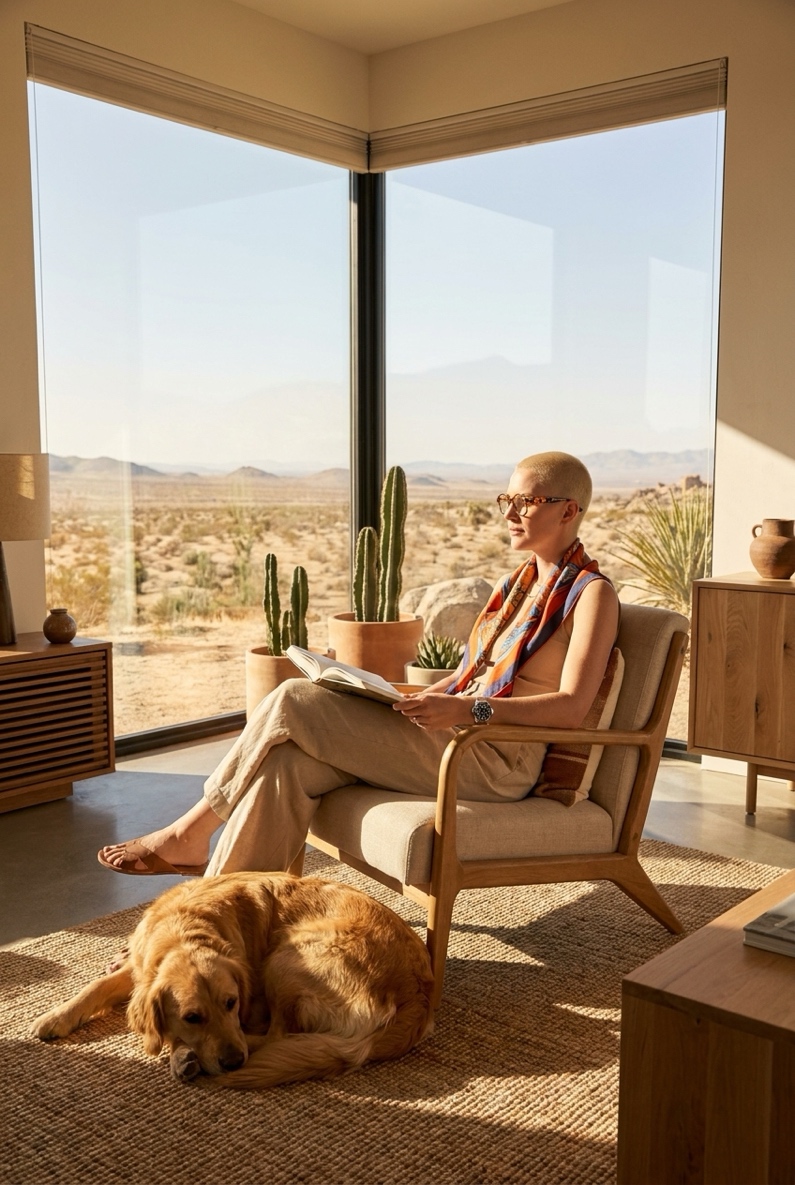} &
\examplejudgement{\textbf{FAIL.} ``The image appears realistic at first glance, but a closer inspection reveals significant issues with object blending and physical impossibility, particularly concerning the watch on the woman's wrist and her foot.''} \\

\exampleimage{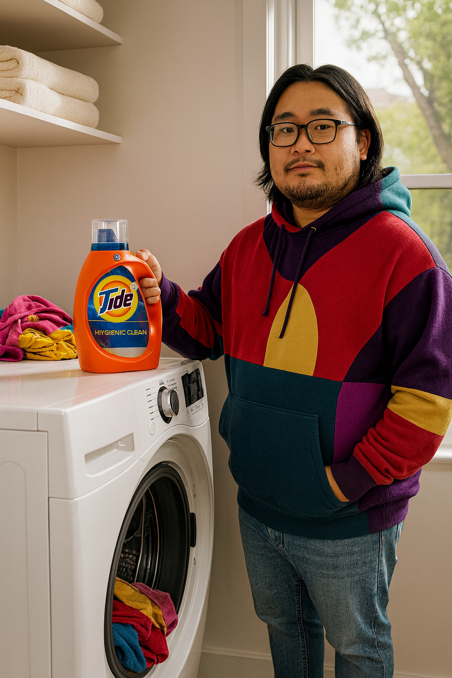} &
\examplejudgement{\textbf{FAIL.} ``The laundry nook scene is generally realistic and well-lit, with plausible geometry, shadows, and anatomy. However, the detergent bottle's front label contains a clear spelling error and multiple packaging inconsistencies, which constitute a major commercial-quality issue. No physics violations are evident beyond these label problems.''} &
\exampleimage{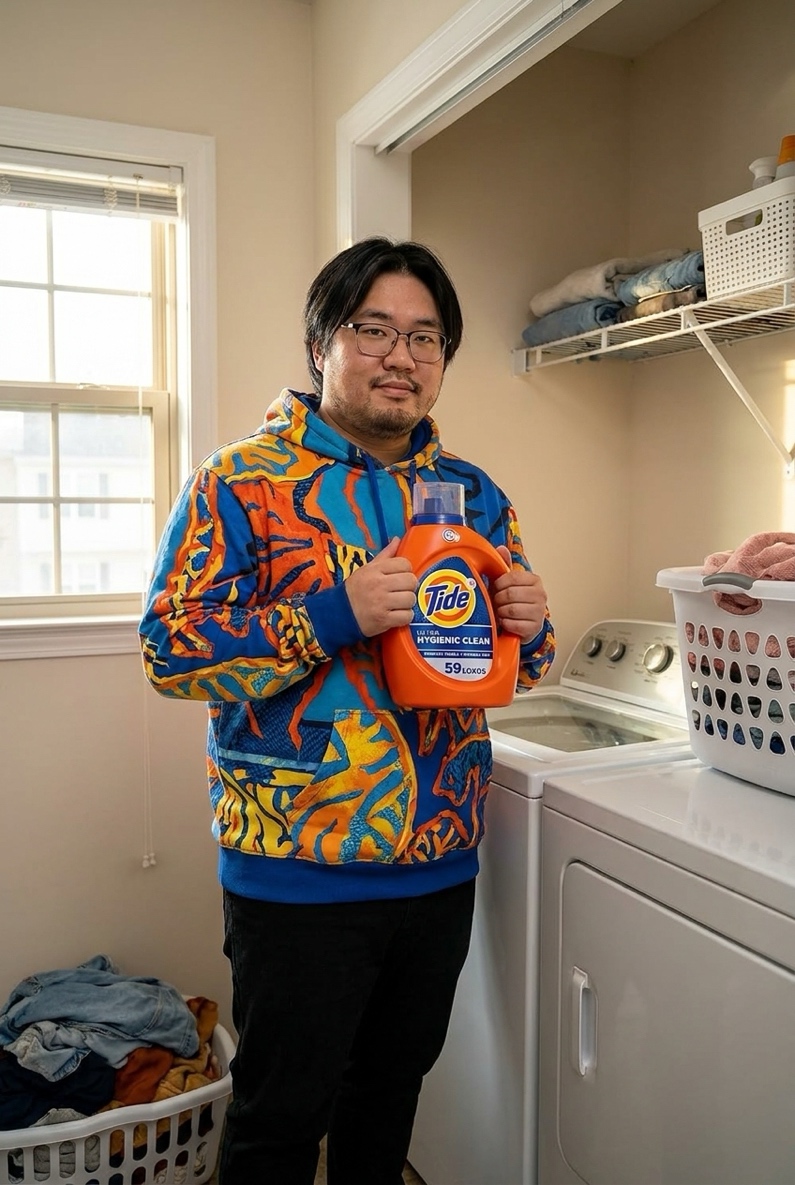} &
\examplejudgement{\textbf{FAIL.} ``The image has a generally photorealistic appearance, but closer inspection reveals several significant anomalies, particularly concerning the man's hands, the text on the detergent bottle, and elements within the background.''} \\

\exampleimage{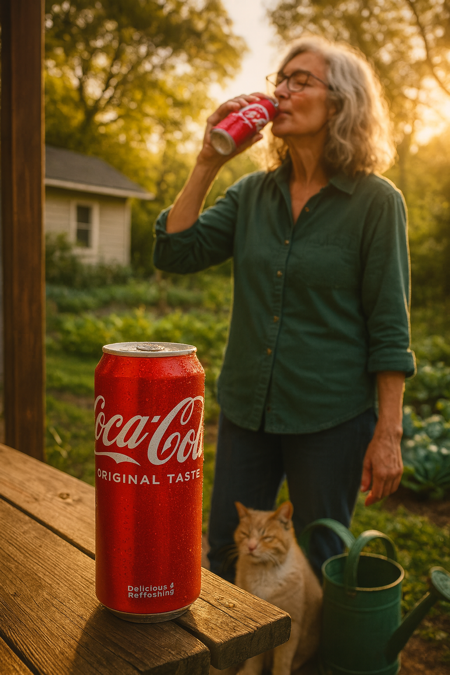} &
\examplejudgement{\textbf{FAIL.} ``Well-composed golden-hour garden scene with convincing depth of field and lighting. Most elements look physically plausible. However, the foreground can have a major packaging/text rendering error, which is unacceptable for commercial use.''} &
\exampleimage{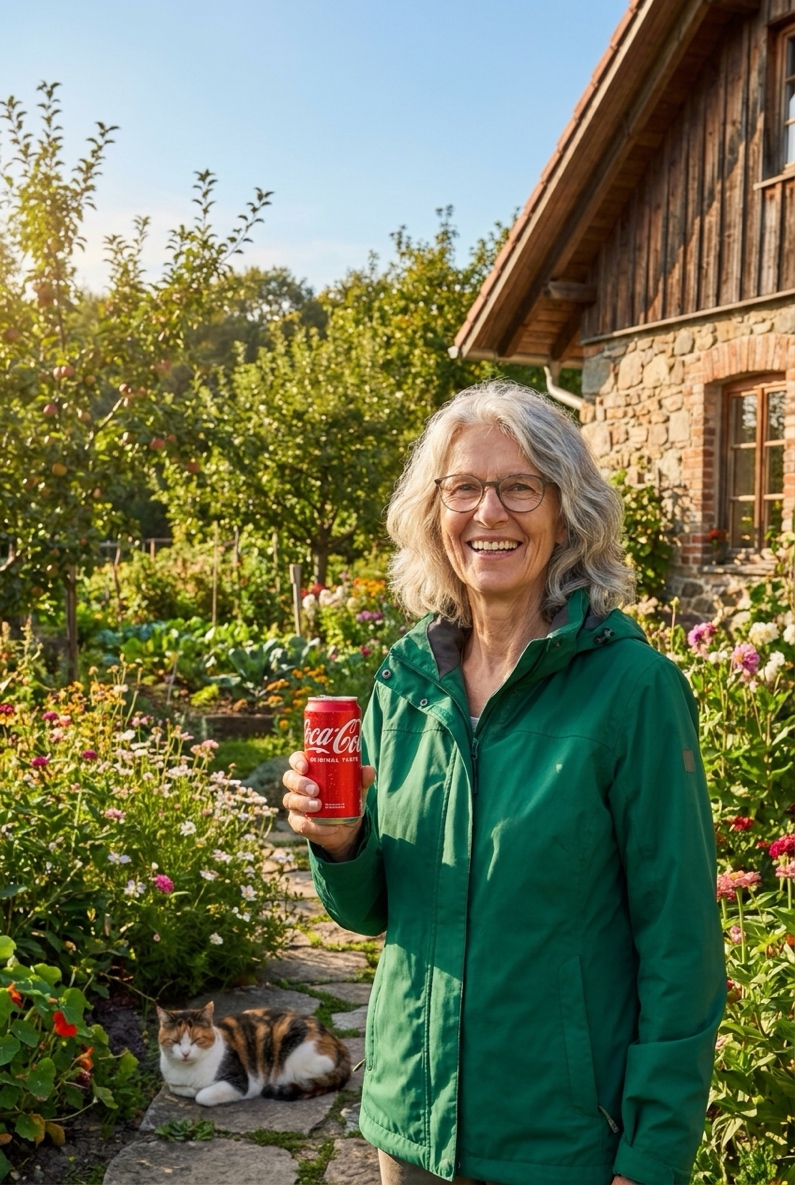} &
\examplejudgement{\textbf{FAIL.} ``The image generally looks realistic at first glance, featuring a woman in a garden. However, closer inspection reveals a severe physical impossibility regarding how the woman is holding the can, along with some structural issues in the background building.''} \\

\bottomrule
\end{tabular}
\endgroup

\caption{Example judge decisions and reasoning for four generated advertisements under the OpenAI GPT and Google Gemini configurations.}
\label{fig:judge_reasoning_examples}
\end{figure*}

\begin{figure*}[tbp]
\centering
\begingroup
\scriptsize
\setlength{\tabcolsep}{2pt}
\renewcommand{\arraystretch}{1.0}
\newcommand{\versionimage}[1]{\begin{minipage}[t]{\linewidth}\vspace{0pt}\centering\includegraphics[width=\linewidth]{#1}\end{minipage}}
\newcommand{\versionjudgement}[2]{\begin{minipage}[t]{\linewidth}\vspace{0pt}\raggedright\textbf{#1.} \textbf{FAIL.} ``#2''\end{minipage}}

\begin{tabular}{@{}>{\centering\arraybackslash}p{0.09\linewidth} p{0.375\linewidth} >{\centering\arraybackslash}p{0.09\linewidth} p{0.375\linewidth}@{}}
\toprule
\multicolumn{2}{@{}c}{\textbf{OpenAI GPT configuration}} &
\multicolumn{2}{c@{}}{\textbf{Google Gemini configuration}} \\
\cmidrule(lr){1-2}\cmidrule(lr){3-4}
\textbf{Image} & \textbf{Judge decision \& reasoning} &
\textbf{Image} & \textbf{Judge decision \& reasoning} \\
\midrule

\versionimage{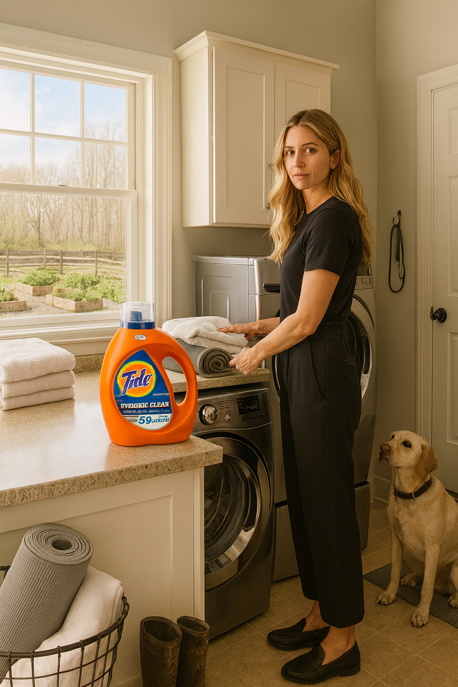} &
\versionjudgement{V0}{The room scene is generally coherent and well lit, but the detergent bottle in the main scene is unrealistically large, and its label is corrupted/ incoherent, and inconsistent with the separate packshot. Additional minor artifacts include oddly merged boots and an incomplete leash. These issues break physical realism and brand integrity.}
&
\versionimage{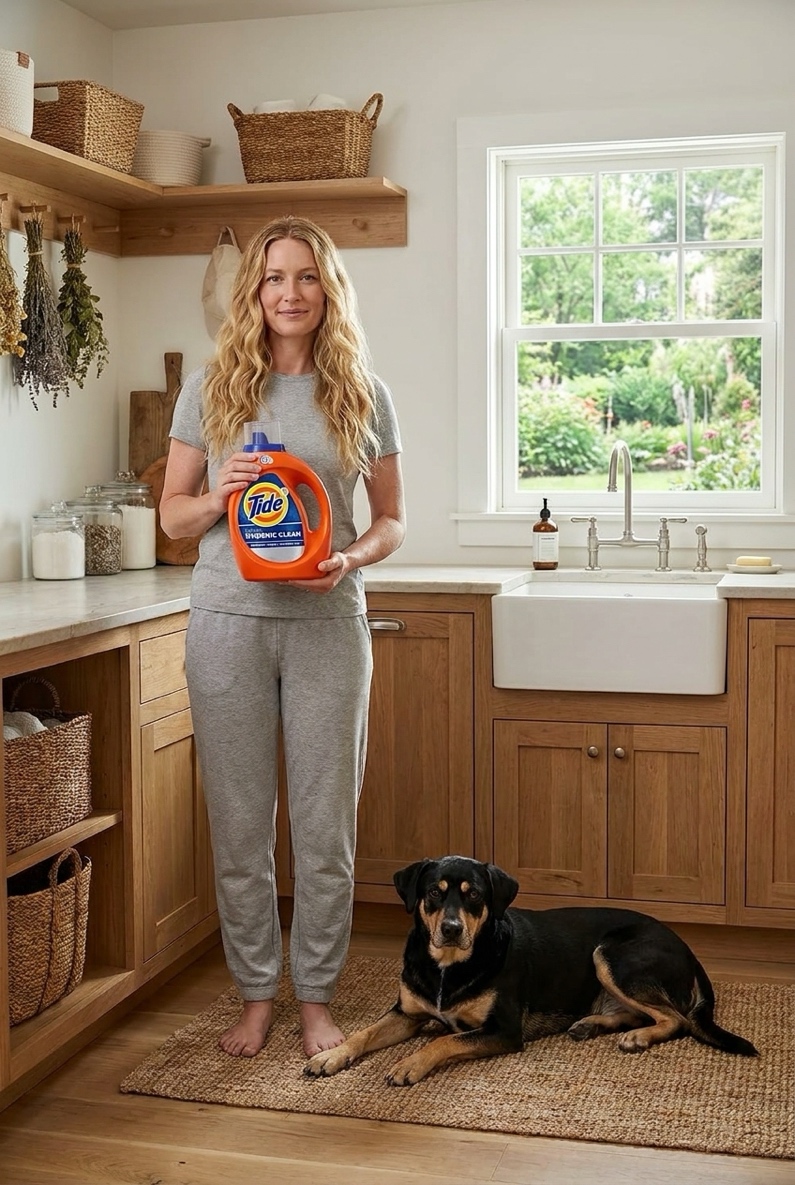} &
\versionjudgement{V0}{The image has significant anomalies, particularly involving the woman's hands, feet, and the dog's anatomy, making it unsuitable for commercial use.} \\

\versionimage{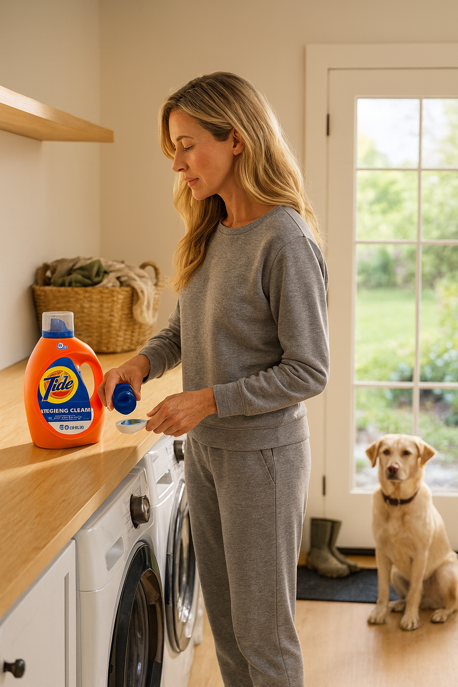} &
\versionjudgement{V1}{Photorealistic, well-lit laundry room scene with accurate lighting and perspective. However, the detergent packaging shows critical inconsistencies: two different caps exist simultaneously (one on the jug and one in the woman's hand), and the label text is corrupted/illegible. These are major issues for commercial use.}
&
\versionimage{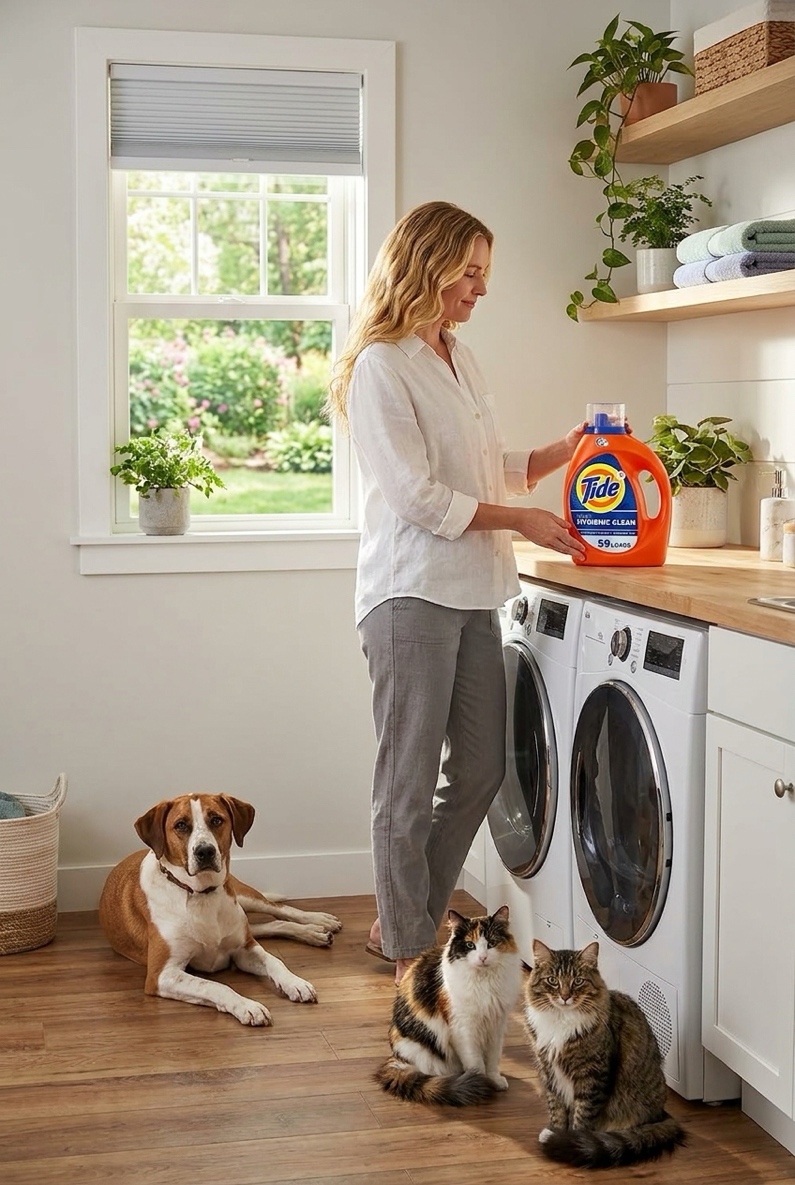} &
\versionjudgement{V1}{The image has multiple major physical and anatomical anomalies, specifically concerning the woman's leg/foot and the structural integrity of the background objects.} \\

\versionimage{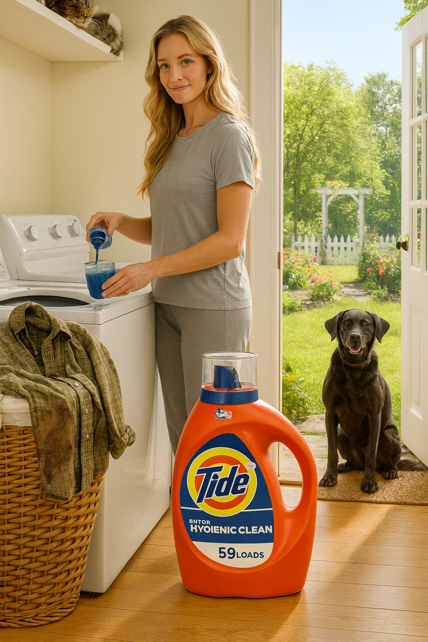} &
\versionjudgement{V2}{Well-lit laundry room scene with garden view looks mostly coherent, but a giant detergent bottle in the foreground breaks real-world scale, and the product label text is garbled. These are major issues for realism and commercial use.}
&
\versionimage{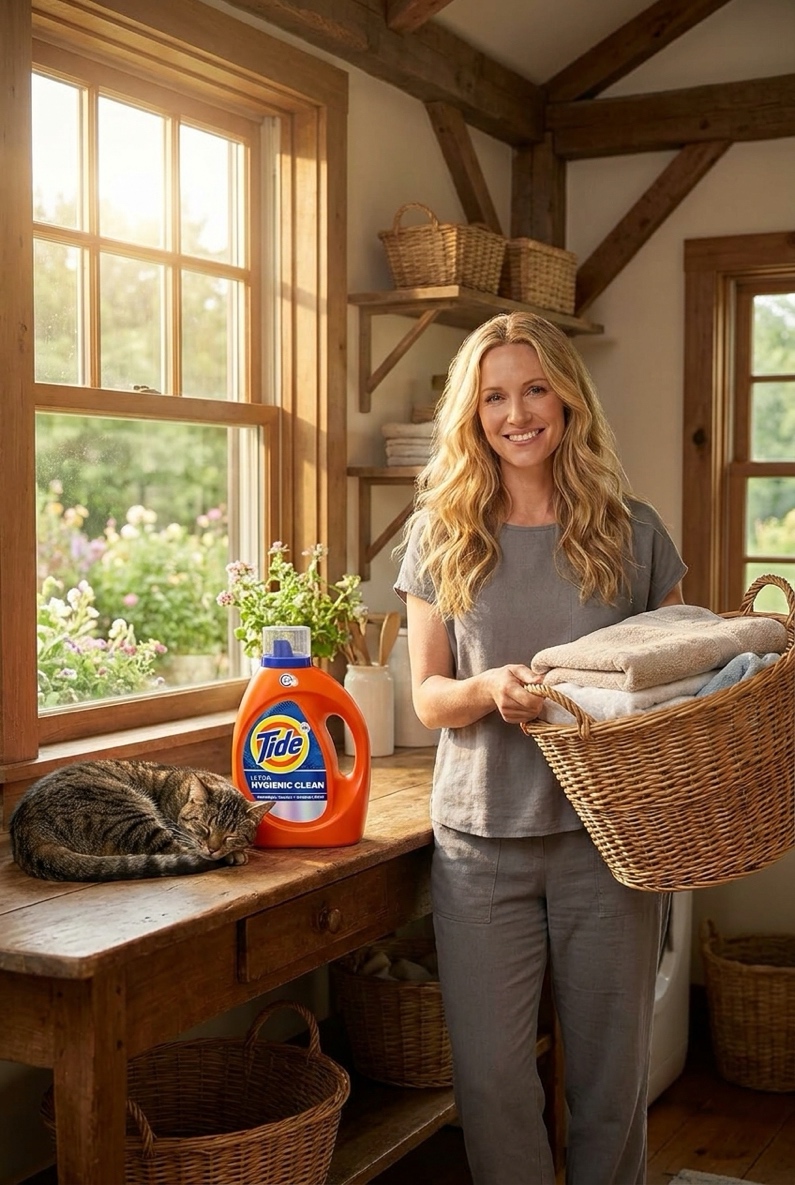} &
\versionjudgement{V2}{The image has a generally realistic appearance at first glance, but several structural and anatomical anomalies reveal its AI-generated nature. The most prominent issues are with the subject's hands and the structure of the window frame.}
\\

\addlinespace[6pt]

\multicolumn{2}{@{}p{0.48\linewidth}}{%
\textbf{SELECTED BEST-OF: V1.} ``All assessed versions fail due to major anomalies, but Version 1 is closest to passing: it lacks the unrealistic scale issue seen in 0 and 2 and has the fewest total anomalies (2 major, 1 minor).''
} &
\multicolumn{2}{p{0.48\linewidth}@{}}{%
\textbf{SELECTED BEST-OF: V2.} ``All versions fail due to major anomalies, but version 2 is selected as it has the fewest major anomalies compared to the severe anatomical deformities (extra limbs, deformed feet/paws) present in v0 and v1.''
} \\
\bottomrule
\end{tabular}
\endgroup

\caption{Compact version comparison under the regeneration policy for the OpenAI GPT and Google Gemini configurations. When no version passes the judge agent, the pipeline selects the best available candidate.}
\label{fig:version_comparison_grid_final_spanning}
\end{figure*}
\section{Discussion} 
\label{sec:discussion}

Our results demonstrate that a fully automated, multi-agent pipeline can generate personalized advertising images from individual customer data at scale, but also reveal systematic challenges that constrain commercial readiness. Comparing the two model configurations separates what is contingent from what is not: artifact prevalence, judge behavior, and product-specific failure modes all shift with the underlying models, whereas text rendering, product fidelity, and automated quality control remain bottlenecks in both.

\subsection{Interpretation of findings}

The visual quality observed across all four product categories indicates that current generative models can produce many professionally plausible images from persona-derived prompts without human intervention. This shows that the step from segment-based personalization of advertisement content to individualized personalization of advertisement content is technically feasible with today's commercial foundation models. At the same time, the comparison of the OpenAI GPT and Google Gemini configurations shows that feasibility should not be equated with model-independent reliability or deployment readiness. The Gemini image generator produces fewer human-labeled artifacts in every image set, and its artifact prevalence is statistically indistinguishable across the electric vehicle, luxury watch, and soft drink. Its judge, however, passes only $40\%$ of final images against $79\%$ for OpenAI GPT. Generation quality and judge strictness therefore vary independently and must be assessed separately.
The effective mapping of personal information from customer profiles to visual outputs (\Cref{subsec:qualitative-results}) suggests that the prompt generation agent can successfully translate persona attributes into the prompts required by image-generation models, and that GenAI-based pipelines can personalize not only \emph{which} product is shown but also \emph{how} it is presented. This extends prior work on generative creative optimization~\citep{czapp2024dynamic,ku2023staging,shilova2023adbooster, smolinski2023towards,vashishtha2024chaining,xu2025personalized,yang2024new}, which has relied on inpainting alone, single personalization dimensions, or coarser personalization such as segments. Notably, what limits the output is not this persona-to-prompt step but the generation step that follows it. Spatial coherence, physical realism, object permanence, and style control all degrade as scene complexity grows.

\paragraph{The influence of product and persona complexity.}
The variation in pipeline performance across products (\Cref{tab:quantitative_results}) reveals that product complexity is central for image-generation quality, a known challenge in GenAI~\citep{chefer2023attend}. Products with simple visual structure and minimal on-package text perform well, whereas products that require accurate text rendering, realistic depiction of functional mechanical parts, or complex spatial arrangements are subject to multiple modes of failure. As shown in \Cref{sec:results}, laundry detergent exemplifies this. A complex bottle shape, textiles, washing machines with doors, buttons, and compartments, and the rendering of liquids jointly bring current image generators to their limits. The Google Gemini image generator reduces some of the structural issues observed in the OpenAI GPT configuration, especially problems with car and washing machine doors, but it does not eliminate the broader product-complexity problem. It still struggles with text, object interaction, floating objects, dog leashes, and detergent bottle lids during pouring (\Cref{fig:example_images_bad}).
Scene complexity, driven by persona detail, compounds this effect. Persona descriptions containing multiple characters, spouses, children, or pets require the model to satisfy many visual constraints simultaneously, and quality degrades accordingly. Product and persona complexity thus appear to interact rather than to add up independently. The personas that are hardest to render are those combined with a product that already imposes structural and typographic constraints of its own. A universal, product-agnostic generation pipeline is therefore unlikely to deliver consistent quality.

\paragraph{Text rendering as a structural limitation.}
The large gap between $AR$ and $rAR$ observed for products with prominent packaging text (\Cref{tab:quantitative_results}) confirms that text rendering remains one of the most pervasive sources of artifacts. This pattern is strongest for the OpenAI GPT configuration, especially for the soft drink and laundry detergent, but it is also present in the Gemini configuration, where laundry detergent still shows a large gap between $AR$ and $rAR$. Even the regeneration mechanism is not able to reliably resolve this issue: repeated generation attempts reproduce the same category of text errors (\Cref{fig:version_comparison_grid_final_spanning}). This sets text rendering apart from other artifact types, which can be corrected through resampling more consistently. The persistence of text errors across model configurations suggests a structural limitation of current image models~\citep{lu2026easytext, rombach2022high}. Addressing this limitation likely requires decoupling text rendering from image generation.

\paragraph{Limitations in physical realism and spatial reasoning.}
The structural artifacts identified in the qualitative evaluation---missing mechanical components, simultaneously open and closed doors, figures partially embedded in objects (\Cref{subsec:qualitative-results})---point to a fundamental gap in the current image generators' ability to represent three-dimensional scene structure and object properties. These defects involve violations of physical plausibility that require implicit knowledge of how objects are constructed and how they interact spatially. The quantitative results suggest that this limitation is model-dependent. Gemini has fewer human-labeled artifacts overall and appears to reduce some mechanical inconsistencies, but it still produces object-interaction errors such as malformed leashes, floating objects, and products that are not held or poured correctly.
The judge agent adds a second limitation that mirrors the first. Within $I_{\text{first}}$, where passed and failed images differ only by the judge's decision, both judges separate artifact-bearing from clean images on $AR$ ($p<0.001$), but neither does so reliably once text-only defects are excluded: the separation is weak for OpenAI GPT ($rAR$ of $0.21$ vs. $0.33$) and not detectable for Google Gemini ($0.03$ vs. $0.06$). The judges therefore operate largely as text-defect detectors, reproducing the very weakness of the generators they are meant to supervise, and structural implausibility passes their review much as it passes the generator's.
Beyond this shared blind spot, the two configurations err in opposite directions. The OpenAI GPT judge accepts too much: $32\%$ of the images it passes carry an artifact that reviewers identified. The Google Gemini judge rejects too much. It fails $78\%$ of first images although only $23\%$ contain an artifact. This asymmetry, rather than any difference in overall judge quality, is what allows the Gemini configuration to combine the lower artifact rate with the lower pass rate. Best-of selection cannot compensate, because it only ranks an existing candidate pool. When every version carries a persistent defect, the least flawed candidate is still flawed.

\paragraph{Exactness of product representation.}
Across products, the generated advertisements generally present the target items in a recognizable and commercially plausible way, indicating that the pipeline is able to preserve core product identity in many cases. At the same time, the outputs exhibit deviations from the real reference products. For instance, the soft drink is sometimes depicted in standard can format instead of sleek can format, the luxury watch can appear too small within the overall composition to retain product-defining details (compare \Cref{fig:image_grid_celebs}), and the electric vehicle occasionally deviates in its proportions or characteristic design lines. Similarly, the laundry detergent bottle is usually recognizable but does not always exactly match the true form or label layout of the reference product, as can be seen in \Cref{fig:version_comparison_grid_final_spanning}. In the Gemini configuration, an additional concern is that the model sometimes reproduces recognizable celebrity faces. While such deviations may not always render an image unusable, they are highly relevant in advertising contexts, where exact product appearance, brand consistency, and rights-sensitive depiction of persons can be important. Future research should investigate how product and identity fidelity can be improved, for example, through fine-tuning models on brand-specific product imagery.

\subsection{Implications for research and practice}
\label{subsec:implications}

\paragraph{Implications for research.}
This work contributes to the field of AI-based personalized advertising by showing that detailed persona attributes such as demographics, lifestyle, and appearance, propagate into visible differences across generated advertisements, and that future work can therefore move beyond mono-dimensional personalization or simple inpainting toward individual-level personalization at scale. Our results equally delimit what is currently achievable: text rendering, physical realism, and reliable automated quality assessment remain open challenges that require advances in both image generation and LLM-based evaluation.
The variation in pipeline performance across different products and model configurations suggests that future research should focus on product-specific, category-aware, and model-aware generation rather than pursuing a universal solution. Products with complex packaging, prominent text, or intricate spatial arrangements pose fundamentally different challenges than products that can be depicted in open, less constrained scenes. Investigating how product-specific constraints can be encoded into the generation and evaluation stages presents a promising direction. At the same time, the measured divergence between human artifact labels and judge decisions shows that automated evaluation needs calibration in two directions at once. Judges must be made sensitive to non-textual, structural defects, which they currently barely detect, without over-penalizing details that human reviewers consider commercially irrelevant. Aligning judge agents with human relevance judgments is thus a prerequisite for automated quality control.

In addition, rather than restarting the full generation cycle, future systems could implement targeted, feedback-driven refinement~\citep{venkatesh2025crea} or leverage inpainting to correct localized defects without regenerating the entire image.

\paragraph{Implications for practice.}
For marketing practitioners, our results indicate that GenAI-based personalized image generation is feasible today for specific product categories, where products have a simple visual structure and limited on-package text. The comparison of the two configurations also yields that a pass rate measures the judge, not the generator, and the two need not agree. In our evaluation the configuration with the substantially lower pass rate is the one that delivers fewer artifacts, so a pass rate is not a usable proxy for delivered quality. Practitioners should therefore validate generation quality and automated quality checks separately, against human review criteria and brand-specific risk thresholds, rather than relying on model-internal judgments. For products with complex packaging or prominent branding text, the pipeline in its current form requires human review before deployment.
The pipeline's modular architecture---with separable persona construction, prompt generation, image generation, and evaluation stages---supports incremental adoption and model substitution. Organizations can integrate individual components into existing workflows (e.g., using the prompt generation agent to assist human designers or using the judge agent to pre-screen manually created variants) without committing to full automation.
If the remaining technical challenges are solved, the practical gains could be especially relevant for social media platform providers. These platforms sit at the intersection of marketers and customers. They host advertisements, maintain active user bases, and possess rich data infrastructures for targeting and personalization. This position suggests a potential future service model in which platforms move beyond serving as distribution channels and offer AI-generated personalized advertising as an optional capability.

\subsection{Limitations \& future work}
\label{subsec:limitations-future-work}

This study has limitations that bound its conclusions and motivate future research. First, our evaluation relies on two instantiations of the \textsc{AdMan} pipeline. The comparison shows that the pipeline is model-dependent, but it should not be read as an exhaustive benchmark of either provider. Different prompt designs, regeneration policies, judge calibration strategies, or refinement mechanisms may yield different performance profiles. Future work should include systematic ablation studies and optimized model-specific pipelines.
Second, the quantitative evaluation is based on human artifact annotations rather than a representative sample of customers. While expert raters provide consistent and technically grounded assessments, future studies should incorporate customer-facing evaluations, including preference studies, engagement experiments, and assessments of perceived personalization.
Third, the data for our persona creation is collected from a predominantly U.S.-based, English-speaking sample via Prolific, limiting the diversity of the personas tested. The pipeline's ability to generate culturally appropriate and non-stereotypical imagery for more diverse populations remains to be validated. In this context, our celebrity personas, while useful for intuitive personalization assessment, are not representative of the broader population.
Fourth, we evaluate four products selected to span the FCB Grid, but this selection does not exhaustively cover the space of advertising contexts. Products with distinct visual challenges (e.g., food items, digital services) may present additional points of failure or strengths not captured in our evaluation. 
Moreover, our quantitative evaluation focuses on visible artifacts and does not measure product fidelity as a separate construct. 
Future work should complement human artifact labels with product-specific fidelity measures such as optical character recognition accuracy for packaging text, logo fidelity, and CLIP-based product matching.
Fifth, the current pipeline does not incorporate feedback loops from actual customer interactions (e.g., click-through rates, dwell times, or purchase conversions). Evaluating the downstream effectiveness of personalized advertisements represents an important direction for future research.
Sixth, we do not quantify computational cost, latency, or deployment overhead. Regeneration increases the number of model calls, and the practical viability of a system such as \textsc{AdMan} depends on the average number of generations per accepted image, API cost, runtime, and whether images must be generated in real time or can be pre-generated for individuals and cached. These operational metrics should be optimized in future deployment-oriented evaluations.
Seventh, our judge-agent analysis quantifies how far judge decisions agree with human artifact labels within $I_{\text{first}}$, but it does not measure calibration, prompt sensitivity, or judge consistency under repeated evaluation of the same image. Future work should therefore evaluate judge agents as independent components, including repeated assessments, alternative judge prompts, and comparisons with specialized computer-vision or vision-language models.
Finally, the ethical dimensions of personalized advertising deserve more systematic investigation. While our pipeline generates content based on voluntarily provided customer attributes, the scalability of such systems raises questions about customer autonomy, persuasion, privacy, and the use of sensitive or quasi-sensitive attributes such as ethnicity, income, body type, family background, and appearance-related features. Future work should develop governance frameworks, risk taxonomies, consent and opt-out mechanisms, and mitigation strategies for such systems in collaboration with customers, regulators, advertisers, and platform providers.
\section{Conclusion}
\label{sec:conclusion}

This paper presents \textsc{AdMan}, a multi-agent pipeline for the fully automated generation of personalized advertising imagery from individual customer data. The pipeline transforms tabular customer attributes into textual personas, translates these personas into image-generation prompts via a prompt generation agent, produces advertisement images using a text-to-image model conditioned on product reference images, and applies an LLM-based judge agent for automated quality control with bounded regeneration. We evaluate this pipeline with both an OpenAI GPT configuration and a Google Gemini configuration.
Our evaluation---combining a qualitative expert focus group with a quantitative artifact-rate assessment---yields three principal findings. First, the pipeline produces personalized advertisements in which persona attributes are reflected in the generated scenes, demonstrating that the step from segment-level targeting to individualized visual presentation is technically feasible (RQ1). Second, image-generation performance differs substantially between model configurations. Third, the LLM-based judge agent measurably improves output quality by filtering defective images, though its effectiveness varies by product complexity, artifact type, and model configuration (RQ2). When no candidate passes, the judge can select the best available version, but this best-of selection remains limited by the quality of the generated candidate pool.
These findings carry direct implications. For research, they motivate the development of product-specific and model-aware generation strategies, feedback-driven refinement mechanisms, calibrated judge agents, and hybrid evaluation architectures. For practice, they indicate that GenAI-based personalized image generation is feasible today for specific product categories, while products with complex packaging, prominent text, or high brand-fidelity requirements still require human oversight. Overall, \textsc{AdMan} demonstrates the promise of personalized advertising imagery, but also its current challenges.
\section*{Declaration of generative AI and AI-assisted technologies in the writing process}

During the preparation of this work, the authors used ChatGPT, Claude, Gemini and DeepL Write in order to assist with language refinement and clarity. After using these tools, the authors reviewed and edited the content as needed and take full responsibility for the content of the published article.

\bibliographystyle{ACM-Reference-Format}
\bibliography{references}  






\end{document}